%% file: iclr2026_conference.tex
\documentclass{article} % For LaTeX2e
\usepackage{iclr2026_conference,times}
\usepackage{graphicx}
\usepackage{multirow}
\usepackage{booktabs}
\usepackage{makecell}
\usepackage{tabularx}
\usepackage{wrapfig}
\usepackage{amssymb} % for checkmark and xmark
\usepackage{pifont}  % for ding symbols
\usepackage{enumitem}
\usepackage{fvextra}
\usepackage[most]{tcolorbox} % 导言区引入
\usepackage{subcaption} % 在导言区
\tcbuselibrary{listings, skins, breakable}

\definecolor{darkgreen}{rgb}{0,0.8,0}

\newtcolorbox{promptbox}[2][]{%
    colback=gray!10,
    colframe=black!50,
    boxrule=0.5pt,
    arc=2pt,
    breakable,
    title={#2},
    #1
}
\tcbset{
  mydefstyle/.style={
    colback=gray!10,    % 背景色
    colframe=black!50,  % 边框色
    boxrule=0.5pt,      % 边框线宽
    arc=2pt,            % 圆角
    left=4pt, right=4pt, top=4pt, bottom=4pt, % 内边距
    boxsep=0pt
  }
}
\usepackage{xcolor}
\usepackage{color}
\usepackage{colortbl}

\newcommand{\cmark}{\ding{51}}%
\newcommand{\xmark}{\ding{55}}%
\renewcommand{\arraystretch}{0.9}
\newcolumntype{Y}{>{\centering\arraybackslash}X} % 定义居中列类型

\input{math_commands.tex}

\usepackage[hidelinks]{hyperref}
\usepackage{url}

\definecolor{mylightblue}{rgb}{0.8,0.9,1.0}
\definecolor{mylighterblue}{rgb}{0.9,0.95,1.0}

\title{From $<$Answer$>$ to $<$Think$>$: Multidimensional Supervision of Reasoning Process for LLM Optimization}

\author{
Beining Wang \\
Tsinghua University \\
Wechat Search, Tencent Inc \\
\texttt{wbn23@mails.tsinghua.edu.cn} \And
Weihang Su \\
Tsinghua University \And
Hongtao Tian \\
Wechat Search, Tencent Inc \And
Tao Yang \\
Wechat Search, Tencent Inc \And
Yujia Zhou \\
Tsinghua University \And
Ting Yao \\
Wechat Search, Tencent Inc \And
Qingyao Ai\thanks{Corresponding author.} \\
Tsinghua University \\
\texttt{aiqy@tsinghua.edu.cn} \And
Yiqun Liu \\
Tsinghua University
}

\iclrfinalcopy % 
\begin{document}

\maketitle

\begin{abstract}
Improving the multi-step reasoning ability of Large Language Models (LLMs) is a critical yet challenging task. 
The dominant paradigm, outcome-supervised reinforcement learning (RLVR), rewards only correct final answers, often propagating flawed reasoning and suffering from sparse reward signals. While process-level reward models (PRMs) provide denser, step-by-step feedback, they lack generalizability and interpretability, requiring task-specific segmentation of the reasoning process.
To this end, we propose the \textbf{D}imension-level \textbf{R}eward \textbf{M}odel (\textbf{DRM}), a new supervision framework that bridges the gap between these two approaches. DRM evaluates the quality of a reasoning process along three fundamental, complementary, and interpretable dimensions: \textbf{Confidence} for uncertainty calibration, \textbf{Relevance} for semantic alignment, and \textbf{Coherence} for logical consistency. 
Together, these dimensions capture aspects beyond final answer correctness and enable interpretable assessment without requiring ground truth answers.
Experimental results show that DRM provides effective supervision signals, guides the optimization of LLMs and enhances their reasoning ability.
In particular, DRM-supervised training achieves consistent gains on both in-distribution and out-of-distribution open-domain tasks, including mathematics, question answering, code execution, and puzzles.
Our findings demonstrate that multidimensional supervision of the reasoning process can improve the generalized reasoning ability of LLMs beyond the training distribution.
\end{abstract}

\section{Introduction}
\label{introduction}
\footnotetext{All models and datasets used in our experiments are downloaded from the Hugging Face Hub. The code is released at \url{https://github.com/Benson0704/DRM}.}
Enhancing the reasoning ability of Large Language Models (LLMs) to perform complex and multi-step reasoning remains a central challenge in their development~\citep{SurveyReinforcement2025,LargeReasoning2025}. 
The dominant paradigm for enhancement relies on Reinforcement Learning with Verifiable Rewards (RLVR)~\citep{DeepSeekMathPushing2024,yang2024qwen2,ImproveMathematical2024}. RLVR provides supervision at the outcome level, assigning a positive reward only if the final answer is correct.
However, this reward mechanism has fundamental limitations.
First, answer supervision overlooks the quality of the reasoning process~\citep{DAPOOpenSource2025}. 
This often leads to rewarding models for arriving at a \textit{correct answer with flawed reasoning} while penalizing sound logic that contains a minor final error~\citep{CAPOEnhancing2025}.
Second, we observed that rewards in RLVR can become nearly constant when the model is either too powerful or too weak on the training set, thereby offering limited guidance for optimization~\citep{EntropyMechanism2025}.
Process-level Reward Models (PRMs) are designed to address these limitations by supervising intermediate steps~\citep{StopSummation2025,OPENPRMBUILDING2025,ReasonFluxPRMTrajectoryAware2025}. While promising, PRMs introduce their own challenges.
Their process-level supervision requires the reasoning process to be segmented into individual steps~\citep{StepWiserStepwise2025,ReasonFluxPRMTrajectoryAware2025}.
This segmentation is often learned in a task-specific manner, which may hinder generalization to open-domain tasks with ambiguous or overlapping steps~\citep{StepWiserStepwise2025}. Furthermore, unlike the transparent binary signal of RLVR, PRMs often function as black-box evaluators, making it difficult to diagnose or trust their scoring mechanism~\citep{DeepReinforcement2023}.

\begin{figure}[t]
\begin{center}
\includegraphics[width=1.0\linewidth]{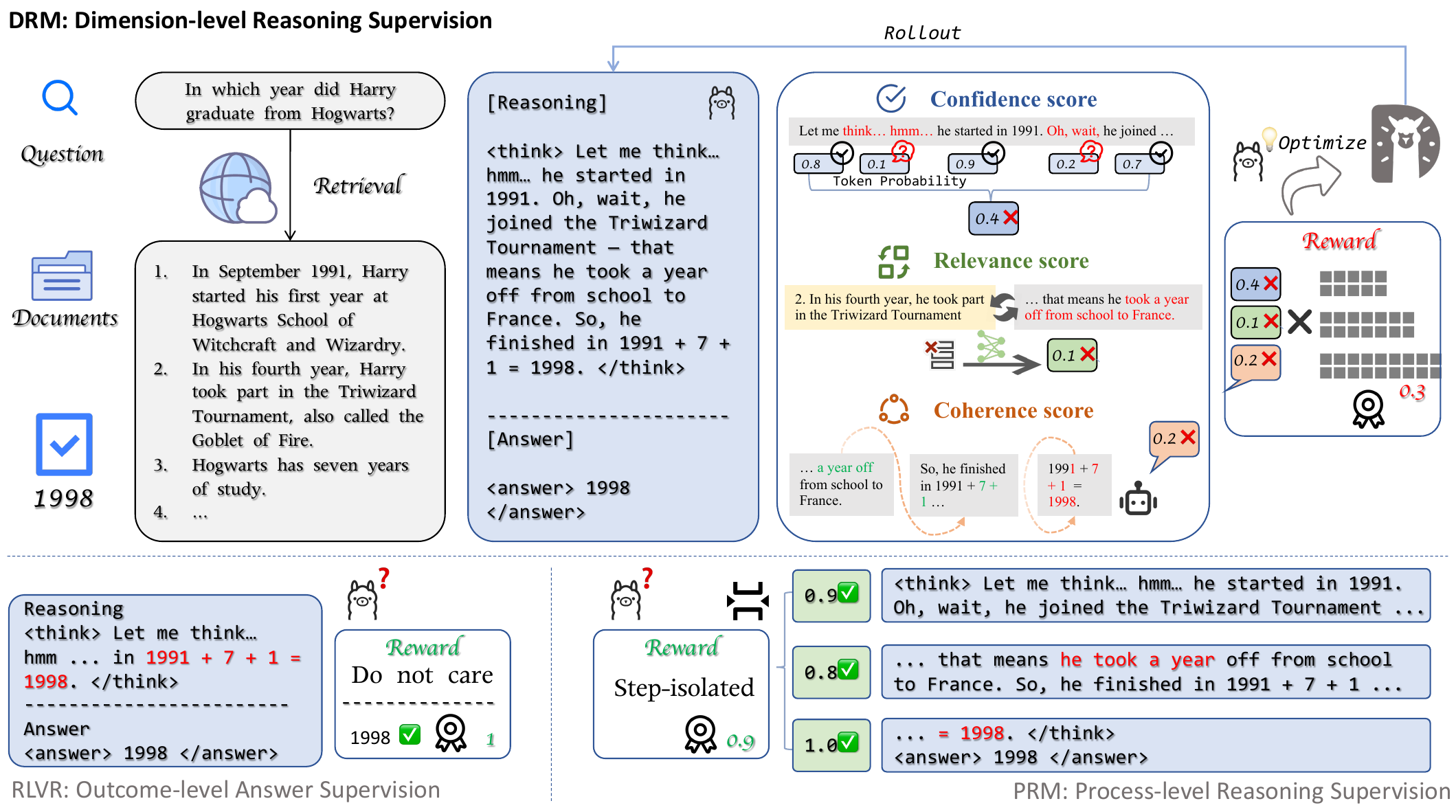}
\end{center}
\caption{
An overview of our multidimensional reasoning supervision framework, illustrated on a RAG task. 
RLVR regards a \textit{correct answer with flawed reasoning} as a positive sample since it focuses solely on the answer.
PRMs also misclassify it because process-level supervision ignores errors across steps when each individual step is correct.
\textbf{DRM} performs dimension-level supervision, detects reasoning flaws, and assigns a reward that reflects the real quality of reasoning process, facilitating further optimization.
}
\label{fig:main}
\end{figure}

To overcome these limitations, we propose a new supervision framework grounded in the key characteristics of a high-quality reasoning process.
Prior work shows that unfaithful content in reasoning process can hinder correct answers~\citep{SurveyReinforcement2025}. 
To detect such content, our framework performs assessment along three complementary dimensions: (1) \textbf{Confidence}, measures whether the reasoning remains faithful to the question and supporting context, directly counters the \textit{flawed reasoning} issue where models hallucinate or deviate; (2) \textbf{Relevance}, evaluates the semantic relatedness and entailment between the reasoning process and the question, the supporting context and the final answer, enabling the detection of deviations from the given information; and (3) \textbf{Coherence}, penalizes self-contradictory statements by the logical consistency of the reasoning process.
\begin{wraptable}{r}{0.49\textwidth}
\centering
\scriptsize

\caption{Comparison of supervision approaches.}
\label{tab:comparison}
\begin{tabular}{lccc}
\toprule
\textbf{Property} & RLVR  & PRM & \textbf{DRM}  \\
\midrule
Supervision level & Outcome  & Process & Dimension\\
Supervision target & Answer  & Reasoning & Reasoning\\
Dense signal & \textcolor{red}{\xmark}  & \textcolor{darkgreen}{\cmark} & \textcolor{darkgreen}{\cmark}\\
Generalization & \textcolor{darkgreen}{\cmark}  & \textcolor{red}{\xmark} & \textcolor{darkgreen}{\cmark}\\
Interpretability & \textcolor{darkgreen}{\cmark}  & \textcolor{red}{\xmark} & \textcolor{darkgreen}{\cmark}\\
Ground truth free & \textcolor{red}{\xmark}  & \textcolor{darkgreen}{\cmark} & \textcolor{darkgreen}{\cmark}\\
\bottomrule
\end{tabular}
\end{wraptable}
Figure~\ref{fig:main} illustrates how our framework assesses the quality of the reasoning process as a \textbf{D}imension-level \textbf{R}eward \textbf{M}odel (\textbf{DRM}) and addresses the limitations of both RLVR and PRMs.
Table~\ref{tab:comparison} summarizes the key properties of the three supervision approaches. By providing a dense, reasoning-aware reward signal without requiring task-specific ground truth answers, DRM overcomes the key limitations of RLVR. Simultaneously, it avoids the task-specific segmentation required by PRMs and offers superior interpretability by scoring reasoning along explicit, diagnosable dimensions.

We demonstrate the effectiveness of DRM-based supervision in both off-policy selection and on-policy training paradigms. The trained models are evaluated on challenging open-domain benchmarks. Our results show that DRM-supervised models perform competitively on both in-distribution and out-of-distribution tasks, indicating stronger generalization than answer-supervised counterparts. For \textsc{Llama-3.1-8B-Instruct}~\citep{Llama32024}, our method achieves performance gains on \textsc{Math500} ($+8.8$, mathematics)~\citep{TrainingVerifiers2021}, \textsc{2Wiki\_RAG} ($+8.7$, multi-hop QA)~\citep{ConstructingMultihop2020} and \textsc{Cruxeval} ($+7.1$, code execution)~\citep{gu2024Cruxeval}. This improvement trend is consistently observed across different models, which unequivocally demonstrates the superiority and generality of DRM supervision. Qualitative analysis and case studies show that DRM mitigates \textit{the correct answer with flawed reasoning} issue common in answer supervision. Our results indicate that multidimensional reasoning supervision enhances the reasoning ability of LLMs and their performance on out-of-distribution tasks.

\section{Methodology: Multidimensional Reasoning Supervision}
\label{methodology}

\textbf{\textit{Task Definition}.} \quad
Formally, let $I$ denote the user input and $O$ the model output. We decompose $O$ into a reasoning process $R$ and an answer $A$. In open-domain scenarios, $I$ often contains more than just the question $Q$. For example, in Retrieval-Augmented Generation (RAG) tasks, $I$ additionally includes retrieved documents, while in preference tasks, $I$ may consist of two candidate responses for the model to compare. Let $D$ denote the additional information accompanying $Q$. Consequently, the input–output structure of the model can be denoted by a quadruple: $(Q, D, R, A)$. In most tasks, the performance of the model is evaluated primarily based on the quality of $A$.

Prior work shows that LLMs sometimes generate unsupported statements during reasoning, which can hinder the production of correct answers~\citep{SurveyReinforcement2025,LargeReasoning2025}. To address this issue, models are expected to produce faithful reasoning that avoids unsupported claims. In particular, they should produce decisive output, especially for the final answer. Furthermore, the reasoning process should be grounded in the provided input and exhibit internal consistency throughout. These properties support both the production of correct answers and the interpretability of reasoning process. We categorize these properties into three dimensions that a high-quality reasoning process should satisfy:  \textbf{Confidence}, \textbf{Relevance} and \textbf{Coherence}. Table~\ref{tab:Reasoning Assessment Dimensions} summarizes their definitions and implementation and the rationale for each is discussed in the following. 

\begin{table*}[t]
\caption{Reasoning assessment dimensions, following the $(Q,D,R,A)$ quadruple format.}
\label{tab:Reasoning Assessment Dimensions}
\centering
\begin{tabularx}{1.02\linewidth}{p{1.7cm} X p{6.7cm}}
\toprule
\textbf{Dimension} & \textbf{Description} & \textbf{Implementation} \\
\midrule
\textbf{Confidence}

$\mathrm{score}^{Conf}$
& Self-assessed certainty of generated $R$ and $A$ from intrinsic signals.& $\mathrm{score}^{Conf}_R=\frac{1}{|R|}\sum\log p$, for all tokens in $R$.

$\mathrm{score}^{Conf}_A=\sum\log p$, for all tokens in $A$.

$\mathrm{score}^{Conf}=\mathrm{score}^{Conf}_R+\mathrm{score}^{Conf}_A$.
\\
\midrule
\textbf{Relevance}

$\mathrm{score}^{Rel}$
& Evaluates whether $R$ is contextually appropriate and semantically aligned with $Q$, $D$ and $A$. & $R \leftarrow Q$: Measured by NLI entailment.

$R \leftrightarrow D$: Measured by semantic relevance.

$R \rightarrow A$: Measured by NLI entailment. 

\\
\midrule
\textbf{Coherence} 

$\mathrm{score}^{Coh}$
& Evaluates logical consistency, fluency and overall quality of $R$. & Evaluated by an external ORM. \\
\bottomrule
\end{tabularx}
\end{table*}

\textbf{Confidence.} \quad This dimension evaluates whether the models are certain about their output. Inspired by prior work on self-confidence evaluation in reasoning models, we compute the average log-probability of tokens in $R$~\citep{PiCSARProbabilistic2025} to avoid penalizing exploratory reasoning processes.
For $A$, we compute the sum of log-probability instead to encourage decisive and confident outputs. 
The final confidence score is calculated as the sum of these two components.\\
\textbf{Relevance.} \quad This dimension assesses whether $R$ maintains necessary relationships with other components $Q$, $D$ and $A$: (1) $Q \rightarrow R$ should hold via Natural Language Inference (NLI) entailment, ensuring $R$ contributes to answering $Q$; (2) $R \leftrightarrow D$ should exhibit high semantic relevance, ensuring $R$ is grounded in the additional information $D$; and (3) $R \rightarrow A$ should also hold via NLI entailment, ensuring $R$ logically leads to $A$. Specifically, we compute the relevance score by framing it as a ranking task: we rank the reasoning process using three distinct metrics, each corresponding to one of the relationships defined earlier, and then combine these scores to obtain the final score.\\
\textbf{Coherence.} \quad This dimension evaluates the text quality ofthe  reasoning process, with attention to coherence and logical consistency. We treat $R$ as the output of a text generation task with the input of $Q,D$. To assess its logical consistency, fluency, and overall textual quality,  we use an external Outcome-level Reward Model (ORM) in the text-quality evaluation. This captures another dimension of reasoning quality that is not directly reflected in confidence or relevance.

Overall, by jointly evaluating the reasoning process along \textbf{Confidence}, \textbf{Relevance} and \textbf{Coherence}, our framework explicitly decomposes assessment into complementary dimensions. As illustrated in Figure~\ref{fig:main}, DRM assesses reasoning quality along three distinct dimensions with each grounded in measurable scores. We compute the DRM reward by a weighted sum of the dimensional scores: 
\begin{equation*}
R^{DRM}_i = \mathrm{score}_i 
= \sum_{D} w^D \,\widetilde{\mathrm{score}}^D_i, 
\quad D \in \{\mathrm{Conf}, \mathrm{Rel}, \mathrm{Coh}\},
\end{equation*}
where $\widetilde{\mathrm{score}}^D_i$ is the component $\mathrm{score}^D_i$ 
after being individually normalized within its group to mitigate scale differences. This produces a dense reward that serves as a direct supervision signal. The weights are determined via a grid search on the validation set. This design inherently avoids the binary sparse reward issue of RLVR and reflects the quality of the reasoning process. DRM replaces stepwise scoring with dimension‑wise assessment and eliminates the need for task‑specific step segmentation in PRMs. Owing to its dimensional nature, DRM inherently provides more interpretable feedback. Moreover, it can distinguish among multiple reasoning processes by their quality, regardless of answer correctness. As DRM addresses the evaluation limitations of RLVR and PRM, we investigate whether its reward can serve as an effective supervision signal for LLM optimization. In off-policy optimization, training sets are constructed under the guidance of a supervision signal. $R^{DRM}_i$ can serve this role by capturing the reasoning quality of each sample, thereby facilitating training set construction. We adopt DPO, and its optimization objective is formulated as follows:
\begin{equation*}
\begin{aligned}
\mathcal{L}_{\text{DPO}}(\theta) 
= - \mathbb{E}_{(I, O^{+}, O^{-})} 
\left[ 
\log \sigma \left( 
\beta \log \frac{\pi_{\theta}(O^{+} \mid I)}{\pi_{\text{ref}}(O^{+} \mid I)}
- \beta \log \frac{\pi_{\theta}(O^{-} \mid I)}{\pi_{\text{ref}}(O^{-} \mid I)}
\right) 
\right],\\
O^+ = \arg\max_{o \in O} R^{DRM}_o,\, O^- = \arg\min_{o \in O} R^{DRM}_o,
\end{aligned}
\end{equation*}
where $\sigma(\cdot)$ is the sigmoid function and $\beta > 0$ controls the sharpness of preference.
In on-policy optimization, DRM can serve as a standalone supervision reward signal, or be integrated with other supervision signals. Specifically, we compute an additional DRM advantage $\hat{A}^{DRM}_{i,t}$
from $R^{DRM}_i$, which denotes the DRM reward for sample $i$. 
We then add this DRM advantage to the native GRPO advantage $\hat{A}_{i,t}$ obtained from RLVR rewards, yielding our optimization objective (for mathematical details, please refer to Appendix~\ref{app:GRPO}):

\begin{equation*}
\begin{aligned}
\mathcal{J}_{\mathrm{GRPO}}(\theta) 
=& \mathbb{E}_{q, \{o_i\}} \frac{1}{G} \sum_{i=1}^G \frac{1}{|o_i|} \sum_{t=1}^{|o_i|} \Bigg\{  
\min \big[ r_{i,t}(\theta)A_{i,t},\; \mathrm{clip}(r_{i,t}(\theta), 1-\varepsilon, 1+\varepsilon) A_{i,t} \big] 
\\&- \beta\,\mathbb{D}_{\mathrm{KL}}\big[\pi_\theta \,\|\, \pi_{\mathrm{ref}}\big] \Bigg\}, \quad A_{i,t} =
\begin{cases}
\hat{A}_{i,t}, & \text{RLVR}, \\
\hat{A}^{DRM}_{i,t}, & \text{DRM}, \\
\hat{A}_{i,t} + \hat{A}^{DRM}_{i,t}, & \text{Combination of RLVR and DRM},
\end{cases}
\end{aligned}
\end{equation*}

where $r_{i,t}(\theta) = \frac{\pi_\theta(o_{i,t} \mid q, o_{i,<t})}{\pi_{\theta_{\mathrm{old}}}(o_{i,t} \mid q, o_{i,<t})}$ is the token-level probability ratio and $\beta$ controls the KL penalty strength with respect to a reference policy $\pi_{\mathrm{ref}}$. DRM can be employed either as a standalone signal or integrated with the RLVR supervision signal.

\section{Experiments}
\label{Experiments}
Following a rigorous experimental paradigm, we formulate a set of research questions to evaluate whether DRM supervision can improve the model’s reasoning ability. The empirical results presented in this section affirmatively answer all of these questions.

\textbf{RQ1:} \quad \textit{Can assessment on reasoning process reliably determine the final answer correctness?}\\
\textbf{RQ2:} \quad \textit{Can the DRM reward signal be learned and used by models to improve reasoning ability?}\\
\textbf{RQ3:} \quad \textit{Can DRM supervision better guide training and outperform RLVR?}\\
\textbf{RQ4:} \quad \textit{Can combining RLVR supervision with DRM supervision lead to further improvements?} \\
\vspace{-1.5em}
\subsection{Experimental Setup}
\label{Experimental Setup}

\textbf{Models.} \quad
We evaluate our method on three representative models: a model lacking inherent reasoning ability \textsc{Llama-3.1-8B-Instruct}~\citep{Llama32024}, a reasoning model \textsc{R1-distil-LLaMA8B}~\citep{DeepSeekR1Incentivizing2025}, 
and a hybrid reasoning model \textsc{Qwen3-8B}~\citep{Qwen3Technical2025}. 
We employ \textsc{Qwen3-8B-reranker}~\citep{Qwen3Embedding2025} as the relevance judge 
and \textsc{Llama-3.3-Nemotron-70B-Reward-Multilingual}~\citep{HelpSteer3PreferenceOpen2025} as the coherence judge.

\textbf{Datasets.} \quad
We evaluate our method on a diverse set of open-domain tasks, including four \textbf{Code} benchmarks, two \textbf{Preference} benchmarks, four \textbf{Math} benchmarks, two \textbf{Scientific QA} benchmarks, three \textbf{Logical Reasoning} benchmarks and two \textbf{Question Answering} benchmarks along with their RAG variants provided by FlashRAG~\citep{FlashRAG}. 
For math tasks, we  use \textbf{\textsc{Math-Verify}}~\citep{math-verify} for automatic solution verification and \textbf{exact match} for all other tasks.\footnote{The main paper only reports results on RewardBench~2; results for HotpotQA with RAG are provided in Appendix~\ref{app:result}.}

\begin{table*}[t]
\small
\centering
\caption{Answer correctness (\%) of reasoning supervision approaches on RewardBench~2. (0.1,0.2,0.7) means weights for Confidence, Relevance and Coherence are 0.1, 0.2, 0.7, respectively. The highest result in each column is in \textbf{bold}.}
\label{tab:reasoning guides answer}
\begin{tabularx}{\linewidth}{lYYYYYY}
\toprule
\multirow{2}{*}{\textbf{Model}} & \multirow{2}{*}{\textbf{Random}} & \multirow{2}{*}{\textbf{Confidence}} & \multirow{2}{*}{\textbf{Relevance}} & \multirow{2}{*}{\textbf{Coherence}} & \textbf{Weighted Equally}
& \textbf{Weighted} 

(0.1,0.2,0.7)
\\
\midrule
LLaMA3.1-8B-Instruct         & 67.17 & 65.44 & 72.32 & 78.55 & 77.45 & \textbf{78.57} \\
R1-Distil-Llama8B   & 63.46 & 63.10 & 66.76 & \textbf{76.35} & 75.11 & 76.16 \\
Qwen3-8B            & 84.87 & 83.20 & 85.10 & 85.54 & 85.01 & \textbf{85.65} \\
\bottomrule
\end{tabularx}

\end{table*}

\subsection{Evaluating Whether DRM Guides Correct Answers}
\label{validating}

To address \textbf{RQ1}, we conduct a comparative evaluation in which models are prompted to generate a reasoning process followed by a final answer. For each instance, we select the reasoning process with the highest DRM reward and evaluate the correctness of the corresponding answer. This correctness directly reflects the effectiveness of the DRM reward signal. We then compare this accuracy with two types of baselines: a baseline obtained via uniform sampling of reasoning processes, which reflects the model’s native performance in the absence of explicit supervision signals, and baselines using each individual DRM dimension (\textbf{Confidence}, \textbf{Relevance}, or \textbf{Coherence}) in isolation, which allows us to assess the contribution of each signal separately.

As shown in Table~\ref{tab:reasoning guides answer}, DRM consistently achieves higher accuracy than the sampling baseline. While using the \textbf{Confidence} score alone slightly reduces accuracy, combining it with \textbf{Relevance} and \textbf{Coherence} improves performance, indicating that these dimensions capture complementary aspects of reasoning quality. We determine the combination weights via grid search on the validation set and fix them for all subsequent experiments. Additional experimental results supporting this choice are provided in Appendix~\ref{app:result}. These findings demonstrate that jointly combining the three dimensions enhances the reliability of reasoning assessment. Overall, the results provide consistent evidence that DRM effectively identifies superior reasoning processes that achieve higher answer correctness, which is desirable for most tasks as their evaluation primarily depends on answer correctness.

\subsection{Assessing the Effectiveness of DRM Supervision}
\label{offpolicy}
This section focuses on \textbf{RQ2} and \textbf{RQ3}. We conduct off‑policy reinforcement learning using DPO with Supervised Fine-Tuning (SFT) loss (for mathematical details, please refer to Appendix~\ref{app:DPOw/SFT}). We construct separate training sets based on different supervision signals. Specifically, DRM rewards serve as reasoning supervision signals, guiding the selection of samples with higher reasoning quality, while RLVR rewards serve as answer supervision signals, selecting samples based on answer correctness.
For each instance in RewardBench~2, we prompt the model to generate $20$ samples containing step‑by‑step reasoning and final answers. These samples are scored and selected according to the respective supervision signal to form preference pairs, as described below. 

\begin{tcolorbox}[mydefstyle]
\textbf{Training Set Construction.}

Let $x$ denote a sample from set $X$, where all samples in $X$ are generated from the same instance. Each sample is associated with a correctness label $\mathrm{answer}_x \in \{\mathrm{True}, \mathrm{False}\}$ and a reasoning quality $\mathrm{score}_x$. The positive set $X^+$ and negative set $X^-$ are defined according to a \textbf{\textsc{SUBSET}} rule and preference pairs are selected according to a \textbf{\textsc{SUPERVISION}} method. Once these two components are specified, the resulting training set is uniquely determined.

\textbf{\textsc{SUBSET}}:\\
\hspace*{2em}\textbf{\textsc{any}}: $X^+ = X^- = X$.\\
\hspace*{2em}\textbf{\textsc{T+T}}: $X^+ = X^- = \{x \mid \mathrm{answer}_x = \mathrm{True}, x \in X\}$.\\
\hspace*{2em}\textbf{\textsc{T+F}}: $X^+ = \{x \mid \mathrm{answer}_x = \mathrm{True}, x \in X\}$,\quad$X^- = \{x \mid \mathrm{answer}_x = \mathrm{False}, x \in X\}$.\\
\hspace*{2em}\textbf{\textsc{F+F}}: $X^+ = X^- = \{x \mid \mathrm{answer}_x = \mathrm{False}, x \in X\}$.\\\textbf{\textsc{SUPERVISION}} \quad \\
\hspace*{2em}\textbf{\textsc{DRM}}: $\{(x^+,x^-)\space|\space x^+ = \arg\max_{x \in X} \mathrm{score}_x,\, x^- = \arg\min_{x \in X} \mathrm{score}_x$\}\\
\hspace*{2em}\textbf{\textsc{RLVR}}: $\{(x^+,x^-)| x^+=\operatorname{random}(X^+),\, x^-=\operatorname{random}(X^-)$\}

Let \textbf{\textsc{SUPERVISION@SUBSET}} denote a training set construction method.
For example, \textbf{\textsc{DRM@T+F}} indicates that we select a sample with the highest DRM reward and correct answer and pair it with a sample with the lowest DRM reward and wrong answer. It is clear that \textbf{\textsc{DRM@any}} refers to the training set constructed with DRM supervision. In contrast, \textbf{\textsc{RLVR@T+F}} refers to the training set constructed with answer supervision, under the RLVR assumption that samples with the same answer are considered equivalent.

\end{tcolorbox}

We construct separate training sets and train models on each set independently.
The full training details are provided in Appendix~\ref{app:settingoffpolicy}.
As shown in Table~\ref{tab:offpolicyrq},  DRM-supervised  training consistently outperforms RLVR-supervised training, providing evidence in support of both research questions.

\textbf{RQ2.} \quad
To assess whether DRM reward signals can be effectively learned and used to improve reasoning ability, we compare \textbf{\textsc{Native}} and \textbf{\textsc{DRM@any}}. 
Additionally, we include \textbf{\textsc{RLVR@any}} as a control group, in which the training set was constructed randomly. 
In the \textbf{\textsc{DRM@any}} setting, the training set is constructed entirely based on DRM reward signals, without incorporating any information about answer correctness. 
Table~\ref{tab:offpolicyrq} shows that \textbf{\textsc{DRM@any}} achieves higher scores than all other settings, with substantial improvements across all evaluated datasets,
The strong performance on out-of-distribution tasks suggests that the model generalizes well beyond the training distribution.
The results indicate that the proposed DRM supervision can be effectively learned even without answer supervision, i.e., without access to the ground truth answers.

\begin{table*}[t]
\centering
\caption{
Results of controlled comparisons for RQ2 and RQ3. 
We use \textsc{LLaMA3.1-8B-Instruct} as the base model. Results for other models, which exhibit the same trend, are provided in Appendix~\ref{app:resultoffpolicy}. As described in Section~\ref{Experimental Setup}, we use \textsc{math-verify} as the evaluation metric for math tasks and EM for all other tasks, respectively. All models are trained for the same number of steps to ensure a fair comparison.
For each row within a comparison, the highest score is in \textbf{bold}.
}
\label{tab:offpolicyrq}
\renewcommand{\arraystretch}{1.1}
\setlength{\tabcolsep}{3pt}
\scriptsize
\begin{tabular}{ll ccc>{\columncolor{mylightblue}}c c>{\columncolor{mylightblue}}c c>{\columncolor{mylightblue}}c}
\toprule
\multirow{3}{*}{\textbf{Task Domain}} & \multirow{3}{*}{\textbf{Dataset}} 
& \multicolumn{4}{c}{\textbf{For RQ2, RQ3.1}} 
& \multicolumn{4}{c}{\textbf{For RQ3.2}} \\
\cmidrule(lr){3-6} \cmidrule(lr){7-8} \cmidrule(lr){9-10}
& & Native 
& \makecell{\textsc{RLVR}\\\textsc{@any}} 
& \makecell{\textsc{RLVR}\\\textsc{@T+F}} 
& \makecell{\textbf{\textsc{DRM}}\\\textbf{\textsc{@any}}}
& \makecell{\textsc{RLVR}\\\textsc{@T+T}}
& \makecell{\textbf{\textsc{DRM}}\\\textbf{\textsc{@T+T}}}
& \makecell{\textsc{RLVR}\\\textsc{@F+F}}
& \makecell{\textbf{\textsc{DRM}}\\\textbf{\textsc{@F+F}}} \\
\midrule
\multirow{4}{*}{Code}
& CodeMMLU   & 58.8 & 58.8 & 59.5 & \textbf{59.9} & 58.9 & \textbf{59.6} & 59.6 & \textbf{61.3} \\
& CodeScope  & 34.8 & 35.4 & 37.4 & \textbf{41.1} & 36.2 & \textbf{41.0} & 36.6 & \textbf{40.0} \\
& Cruxeval   & 50.4 & 53.5 & 52.6 & \textbf{57.5} & 53.6 & \textbf{56.6} & 53.9 & \textbf{55.9} \\
& Execution-v2 & 38.2 & 40.9 & 43.2 & \textbf{45.3} & 39.2 & \textbf{45.5} & 40.3 & \textbf{46.8} \\
\midrule
\multirow{2}{*}{Preference}
& RM-Bench  & 56.4 & 59.3 & 59.2 & \textbf{61.0} & 60.0 & \textbf{60.3} & 59.7 & \textbf{61.9} \\
& UltraFeedback & 66.6 & 65.6 & 65.4 & \textbf{69.9} & 66.4 & \textbf{67.7} & 64.5 & \textbf{68.8} \\
\midrule
\multirow{4}{*}{Math}
& AIME24    & 4.7 & 4.7 & 4.0 & \textbf{6.0} & 4.7 & \textbf{7.3} & \textbf{4.7} & 4.0 \\
& AMC23     & 22.5 & 23.5 & 23.5 & \textbf{29.5} & 23.0 & \textbf{25.5} & 22.0 & \textbf{26.5} \\
& GSM8K     & 88.8 & 89.0 & 89.5 & \textbf{91.8} & 90.2 & \textbf{91.7} & 88.7 & \textbf{91.7} \\
& Math500   & 39.6 & 41.4 & 43.4 & \textbf{48.4} & 42.0 & \textbf{46.6} & 40.0 & \textbf{48.4} \\
\midrule
\multirow{2}{*}{Scientific QA}
& MMLU-Pro  & 41.9 & 45.3 & 46.4 & \textbf{48.7} & 45.7 & \textbf{48.4} & 46.6 & \textbf{49.0} \\
& GPQA      & 31.3 & 28.8 & 32.8 & \textbf{35.9} & 29.8 & \textbf{30.3} & 29.8 & \textbf{35.4} \\
\midrule
\multirow{3}{*}{Reasoning}
& MuSR    & 48.3 & 49.5 & 49.7 & \textbf{51.7} & 48.3 & \textbf{53.3} & 49.7 & \textbf{51.6} \\
& DROP    & 56.9 & 61.0 & 62.9 & \textbf{63.6} & 60.0 & \textbf{64.4} & 58.5 & \textbf{65.1} \\
& QASC    & 84.4 & 84.0 & 84.2 & \textbf{87.2} & 83.8 & \textbf{87.8} & 83.4 & \textbf{86.2} \\
\midrule
\multirow{2}{*}{QA}
& 2wiki     & 33.8 & 33.2 & 34.6 & \textbf{35.6} & 32.3 & \textbf{32.7} & 30.7 & \textbf{33.4} \\
& HotpotQA  & 29.3 & 29.9 & 30.1 & \textbf{31.8} & 29.3 & \textbf{30.1} & 29.1 & \textbf{29.7} \\
\multirow{2}{*}{QA-RAG}
& 2wiki\_RAG & 31.2 & 32.1 & 35.8 & \textbf{39.9} & 36.6 & \textbf{41.4} & 32.1 & \textbf{43.3} \\
& HotpotQA\_RAG & 28.3 & 28.3 & 32.3 & \textbf{34.5} & 29.3 & \textbf{32.3} & 28.5 & \textbf{33.8} \\
\bottomrule
\end{tabular}
\end{table*}

\textbf{RQ3.} \quad
We compare DRM and RLVR across two key aspects to assess their relative effectiveness:  \\
\textbf{\textit{Performance gain:}} To evaluate the effectiveness of DRM, we compare \textbf{\textsc{RLVR@T+F}} with \textbf{\textsc{DRM@any}} (see Table~\ref{tab:offpolicyrq}). 
This comparison examines whether explicit supervision of reasoning achieves better performance than supervising only the answer.
In this setting, \textbf{\textsc{DRM@any}} consistently achieves higher performance than \textbf{\textsc{RLVR@T+F}}, indicating that training with DRM supervision consistently outperforms RLVR supervision.\\
\textbf{\textit{Overcoming limitations:}} We compare \textbf{\textsc{RLVR@T+T}} with \textbf{\textsc{DRM@T+T}} and \textbf{\textsc{RLVR@F+F}} with \textbf{\textsc{DRM@F+F}} to test whether DRM can still provide supervision when all answers have identical correctness labels, where RLVR cannot produce a preference signal. 
Results show that DRM can distinguish reasoning quality in such case, demonstrating its ability to generate informative supervision and to enhance the model's ability to handle a broader range of scenarios.

Furthermore, we conduct off-policy training and compare it against the baselines as shown in Table~\ref{tab:DPOresult}. 
We evaluate our model against three strong baselines: 
(1) a model trained on the \textbf{\textsc{any}} subset with reasoning supervision signals from \textsc{Skywork-Reward-V2-Llama-3.1-8B}, a powerful ORM, 
(2) \textsc{RLPR}~\citep{RLPRExtrapolating2025} and 
(3) \textsc{Klear}~\citep{KlearReasonerAdvancing2025}. 
Both \textsc{RLPR} and \textsc{Klear} are reasoning‑enhanced models trained using the same backbone architecture as their counterparts in our experiments.
This setup allows us to examine whether our DRM provides more effective and generalizable supervision than existing reasoning‑supervision approaches. We also examines whether DRM-supervised models can outperform models optimized with other methods.
Across most downstream open‑domain tasks, DRM outperforms all three baselines. In particular, it surpasses \textsc{RLPR} and \textsc{Klear} under the same backbone, demonstrating its effectiveness. It also exceeds the performance of the model trained with \textsc{Skywork} supervision, indicating that DRM consistently achieves stronger and more generalizable reasoning ability.
The improvements are consistent across various architectures and tasks, suggesting that DRM is an architecture‑agnostic approach that generalizes well.
Notably, our training relies solely on preference data from RewardBench~2, the same type of data used for training reward models~\citep{OPENPRMBUILDING2025,ComprehensiveSurvey2025}, without access to ground truth answers or task-specific finetuning. 
This highlights the data efficiency of our approach as a single source of preference data leads to broad improvements across open-domain tasks.

\begin{table*}[t]
\centering
\caption{Results of off-policy DPO with SFT loss training. 
\textbf{\textsc{RLPR}} and \textbf{\textsc{Klear}} are baseline models that share the same backbone architectures as their respective counterparts.
\textbf{\textsc{Skywork}} indicates that the model's training set is constructed using \textsc{Skywork} reward model. \textbf{DRM} represents \textbf{\textsc{DRM@any}}. For each row within a model group, the highest score is in \textbf{bold}.}
\label{tab:DPOresult}
\renewcommand{\arraystretch}{1.1}
\setlength{\tabcolsep}{3pt}
\scriptsize
\begin{tabular}{ll ccc>{\columncolor{mylightblue}}c cc>{\columncolor{mylightblue}}c ccc>{\columncolor{mylightblue}}c}
\toprule
\multirow{3}{*}{\textbf{Task Domain}} & \multirow{3}{*}{\textbf{Dataset}} 
& \multicolumn{4}{c}{\textbf{LLaMA3.1-8B-Instruct}} 
& \multicolumn{3}{c}{\textbf{R1-Distil-Llama8B}} 
& \multicolumn{4}{c}{\textbf{Qwen3-8B}} \\
\cmidrule(lr){3-6} \cmidrule(lr){7-9} \cmidrule(lr){10-13}
& & Native & \textsc{RLPR} & \textsc{Skywork} & \textbf{DRM}
  & Native & \textsc{Skywork} &  \textbf{DRM}
  & Native & \textsc{Klear} & \textsc{Skywork} &  \textbf{DRM} \\
\midrule
\multirow{4}{*}{Code} 
& CodeMMLU   & 58.8 & 58.0 & 57.6 & \textbf{59.9} & 59.7 & 62.9 & \textbf{66.3} & 77.9 & 77.4 & 79.3 & \textbf{80.3} \\
& CodeScope  & 34.8 & 38.7 & 39.3 & \textbf{41.1} & 67.4 & 68.2 & \textbf{70.2} & 86.5 & \textbf{88.1} & 86.2 & 87.4 \\
& Cruxeval   & 50.4 & 53.6 & 53.6 & \textbf{57.5} & 71.9 & 77.0 & \textbf{77.2} & 91.6 & 87.2 & 91.9 & \textbf{93.0} \\
& Execution-v2 & 38.2 & 44.7 & 42.8 & \textbf{45.3} & 80.8 & 82.0 & \textbf{86.0} & 98.5 & 95.2 & 97.9 & \textbf{99.0} \\
\midrule
\multirow{2}{*}{Preference} 
& RM-Bench  & 56.4 & 60.2 & 59.8 & \textbf{61.0} & 71.9 & 73.4 & \textbf{74.6} & 85.4 & 83.7 & 85.1 & \textbf{85.6} \\
& UltraFeedback   & 66.6 & 68.5 & 67.0 & \textbf{69.9} & 65.2 & 66.5 & \textbf{66.8} & 71.3 & 68.1 & 72.2 & \textbf{73.2} \\
\midrule
\multirow{4}{*}{Math} 
& AIME24    & 4.7  & \textbf{6.0}  & 4.0  & \textbf{6.0}  & 28.7 & 26.7 & \textbf{33.3} & 38.0 & 40.0 & 38.7 & \textbf{44.7} \\
& AMC23     & 22.5 & 26.0 & 25.5 & \textbf{29.5} & 70.5 & 74.5 & \textbf{75.5} & 72.0 & 75.0 & 76.0 & \textbf{79.0} \\
& GSM8K     & 88.8 & 90.0 & 89.8 & \textbf{91.8} & 66.7 & \textbf{73.7} & 69.2 & 95.6 & 93.8 & 95.8 & \textbf{96.1} \\
& Math500   & 39.6 & 47.2 & 42.6 & \textbf{48.4} & 62.6 & \textbf{65.6} & 63.2 & 73.2 & 68.2 & 72.6 & \textbf{75.6} \\
\midrule
\multirow{2}{*}{Scientific QA} 
& MMLU-Pro  & 41.9 & 36.3 & 46.7 & \textbf{48.7} & 51.5 & 52.8 & \textbf{54.7} & 65.3 & 67.1 & 70.0 & \textbf{71.4} \\
& GPQA      & 31.3 & 30.8 & 33.3 & \textbf{35.9} & 39.9 & 37.4 & \textbf{44.9} & 48.0 & 55.6 & 52.5 & \textbf{58.1} \\
\midrule
\multirow{3}{*}{Reasoning} 
& MuSR    & 48.3 & 48.7 & 49.7 & \textbf{51.7} & 52.6 & 52.8 & \textbf{54.1} & 63.5 & 50.8 & 63.5 & \textbf{65.5} \\
& DROP    & 56.9 & 45.4 & 63.0 & \textbf{63.6} & 50.8 & \textbf{54.5} & 50.2 & 74.7 & 68.8 & 74.2 & \textbf{74.9} \\
& QASC    & 84.4 & 87.0 & 87.1 & \textbf{87.2} & 82.1 & 82.5 & \textbf{84.1} & 94.1 & 93.3 & 93.7 & \textbf{94.2} \\
\midrule
\multirow{2}{*}{QA} 
& 2wiki     & 33.8 & 32.1 & 32.4 & \textbf{35.6} & 26.2 & 29.3 & \textbf{31.6} & 39.8 & 35.9 & 40.0 & \textbf{42.2} \\
& HotpotQA  & 29.3 & 29.9 & 30.4 & \textbf{31.8} & 18.1 & 19.3 & \textbf{19.7} & 29.2 & 19.6 & 29.1 & \textbf{29.4} \\
\multirow{2}{*}{QA-RAG} 
& 2wiki\_RAG & 31.2 & 38.7 & 34.8 & \textbf{39.9} & 36.7 & \textbf{39.2} & 37.9 & 55.7 & 52.2 & 55.8 & \textbf{56.1} \\
& HotpotQA\_RAG & 28.3 & 32.8 & 33.2 & \textbf{34.5} & 27.1 & 26.5 & \textbf{27.3} & 40.5 & 34.3 & 40.3 & \textbf{40.7} \\
\bottomrule
\end{tabular}
\end{table*}

\subsection{Enhancing RLVR with DRM}
\label{onpolicy}

This section addresses \textbf{RQ4}. We conduct on-policy GRPO training on three advantage configurations: answer supervision only, reasoning supervision only and their combination. This setup directly tests whether DRM supervision and integrating DRM rewards into RLVR achieve further gains. The comparison between RLVR and DRM also examines whether the trend observed in off-policy training remains consistent in on-policy stages. GRPO training details are provided in Appendix~\ref{app:settingonpolicy}.

Across most model backbones and representative benchmarks on open-domain tasks, the combined approach performs as well as or better than the best single supervision approach, as shown in Table~\ref{tab:GRPOresult}. 
This trend is also consistently observed in the off-policy setting.
The combination also outperforms RLVR, indicating that incorporating reasoning supervision alongside answer supervision consistently improves performance by guiding intermediate reasoning steps during policy optimization. 
When compared to DRM, the combination yields gains, but shows slight drops in certain reasoning-focused or knowledge-intensive datasets, such as MuSR and GPQA, suggesting that in these cases direct RLVR may interfere with the optimization due to overlooking the reasoning process.
Overall, these findings indicate that integrating answer and reasoning supervision provides stable improvements across diverse open-domain tasks, supporting an affirmative answer to \textbf{RQ4}.

\section{Analysis}
\begin{table*}[t]
\centering
\caption{Results of on-policy GRPO training. \textbf{RLVR} denotes training with answer supervision only. 
\textbf{DRM} denotes training with reasoning supervision only. 
\textbf{Combination} denotes training with their combination. Only representative benchmarks are reported here for brevity, with complete results in Appendix~\ref{app:resultonpolicy}. For each row within a model group, the highest score is in \textbf{bold}.}
\label{tab:GRPOresult}
\renewcommand{\arraystretch}{1.1}
\setlength{\tabcolsep}{3pt}
\scriptsize
\begin{tabular}{ll c>{\columncolor{mylighterblue}}c>{\columncolor{mylightblue}}c c>{\columncolor{mylighterblue}}c>{\columncolor{mylightblue}}c c>{\columncolor{mylighterblue}}c>{\columncolor{mylightblue}}c}
\toprule
\multirow{2}{*}{\textbf{Task Domain}} & \multirow{2}{*}{\textbf{Dataset}} 
& \multicolumn{3}{c}{\textbf{LLaMA3.1-8B-Instruct}} 
& \multicolumn{3}{c}{\textbf{R1-Distil-Llama8B}} 
& \multicolumn{3}{c}{\textbf{Qwen3-8B}} \\
\cmidrule(lr){3-5} \cmidrule(lr){6-8} \cmidrule(lr){9-11}
& & RLVR & DRM & Combination
  & RLVR & DRM & Combination
  & RLVR & DRM & Combination \\
\midrule
\multirow{2}{*}{Code} 
& CodeScope     & 37.2 & 39.4 & \textbf{40.5} & 69.2 & 68.2 & \textbf{70.8} & 87.3 & \textbf{87.7} & 87.5 \\
& Execution-v2  & 44.7 & 42.4 & \textbf{46.4} & 82.3 & 83.5 & \textbf{85.6} & 98.5 & 99.0 & \textbf{99.2} \\
\midrule
\multirow{3}{*}{Math} 
& AIME24        & \textbf{4.7} & \textbf{4.7} & \textbf{4.7} & 29.3 & \textbf{34.7} & 33.3 & 38.0 & \textbf{46.7} & 45.3 \\
& AMC23         & 20.5 & 23.0 & \textbf{24.5} & 70.5 & 77.5 & \textbf{80.5} & 75.0 & \textbf{81.5} & 79.5 \\
& Math500       & 40.8 & 38.0 & \textbf{45.4} & 62.8 & 67.0 & \textbf{67.2} & 73.8 & \textbf{75.8} & \textbf{75.8} \\
\midrule
\multirow{2}{*}{Scientific QA} 
& MMLU-Pro      & 42.3 & 43.2 & \textbf{47.8} & 53.6 & 53.4 & \textbf{54.1} & 63.7 & 68.7 & \textbf{69.1} \\
& GPQA          & 30.8 & 28.8 & \textbf{32.3} & 39.4 & \textbf{43.9} & 42.4 & 43.9 & \textbf{57.6} & 56.6 \\
\midrule
Reasoning 
& MuSR          & 47.6 & \textbf{52.9} & 52.1 & \textbf{53.0} & \textbf{53.0} & 52.9 & 63.0 & 63.2 & \textbf{64.3} \\
\bottomrule
\end{tabular}
\end{table*}

\subsection{Can DRM Identify High-Quality Reasoning Process?}
\label{ablation2}

As introduced in Section~\ref{methodology}, most tasks are evaluated solely based on answer correctness, regardless of the quality of the reasoning process that produced the answer. However, a clear and coherent reasoning process helps users assess and trust the output in interactions with LLMs. This section examines whether DRM can identify truly high-quality reasoning process.
We prompt GPT-4o to determine whether a reasoning process and its corresponding answer constitute a \textit{correct answer with flawed reasoning} in off-policy training sets constructed with two different supervision approaches. In these settings RLVR denotes answer supervision while DRM denotes reasoning supervision.
As shown in Figure~\ref{fig:ablation2}, the number of \textit{correct answer with flawed reasoning} instances decreases substantially across all models when using DRM. These results indicate that DRM prioritizes instances with higher reasoning quality compared to RLVR, confirming that reasoning supervision successfully identifies real high-quality reasoning process associated with completely correct answers.
Together with the experiments addressing \textbf{RQ1} in Section~\ref{validating}, we demonstrate that our multidimensional reasoning supervision not only produces more correct answers but also improves reasoning quality by reducing \textit{correct answer with flawed reasoning}.

\subsection{Ablation Study of Individual Supervision Dimensions}
\label{ablation1}

\begin{figure}[ht]
    \centering
    \begin{subfigure}{0.44\linewidth}
        \centering
        \includegraphics[width=\linewidth]{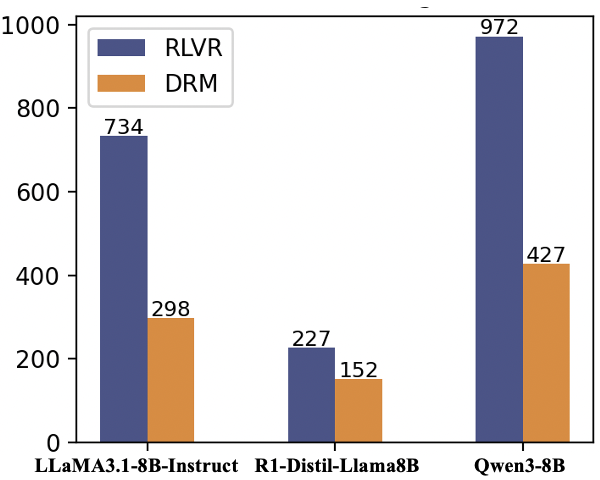}
           \caption{Count of \textit{correct answers with flawed reasoning} as evaluated by GPT-4o. Each training set contains approximately 6{,}000 samples.} 
        \label{fig:ablation2}
    \end{subfigure}
    \hfill
    \begin{subfigure}{0.55\linewidth}
        \centering
        \includegraphics[width=\linewidth]{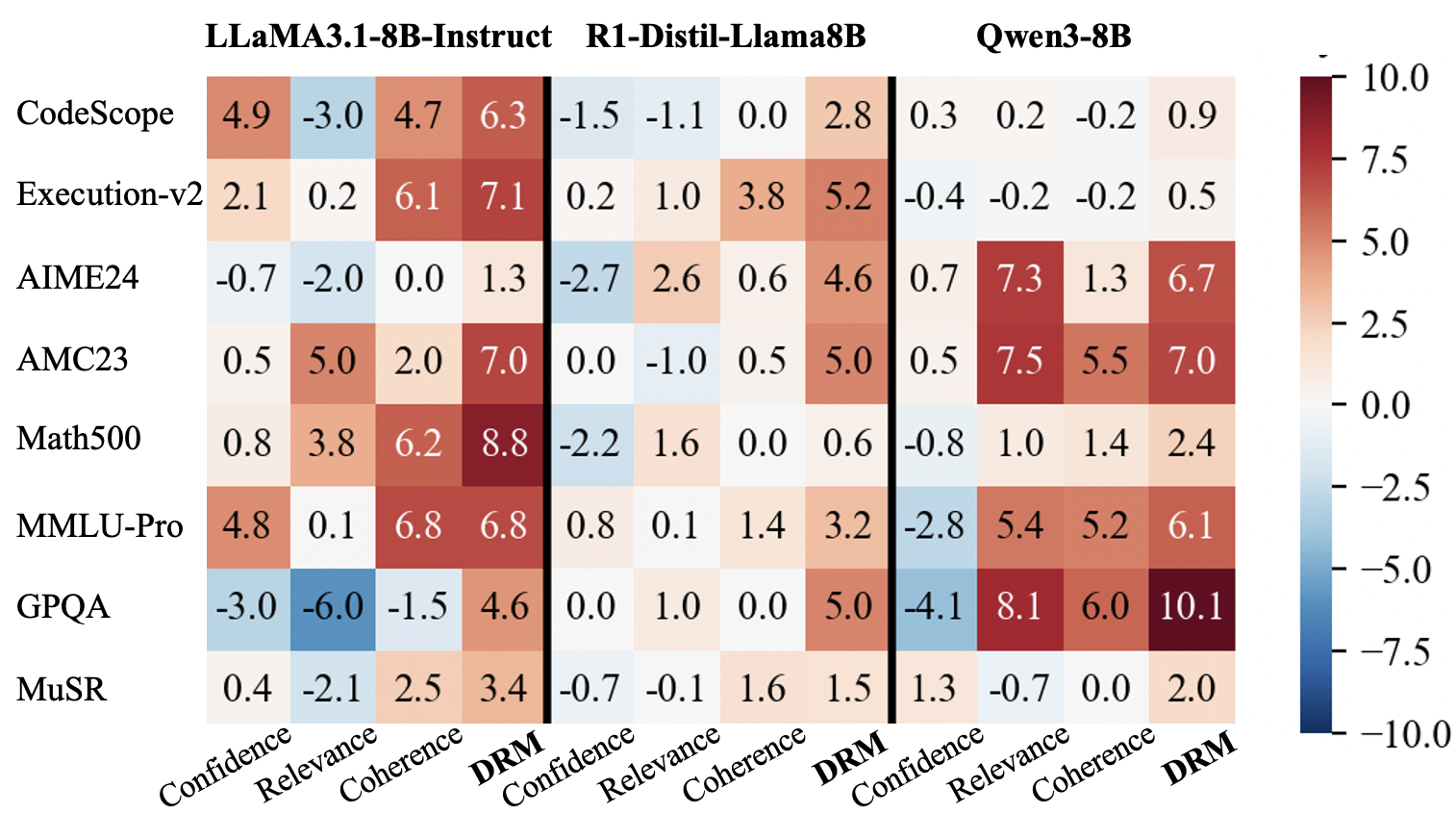}
           \caption{Ablation results of single dimension supervised training. The values in this heatmap indicate the absolute difference relative to the native model. Training pairs are selected from \textbf{\textsc{any}} subset. \textbf{DRM} means training with DRM supervision.} 
        \label{fig:ablation1}
    \end{subfigure}
    \caption{Analysis of DRM supervision effectiveness: (a) reduction in flawed reasoning cases; (b) performance under single dimension supervision.}
    \label{fig:both}
\end{figure}

We conduct an ablation study to examine the effect of each reasoning supervision dimension in isolation. Starting from the native model, we adopt the same off-policy training setting and apply supervision to only one dimension at a time: \textbf{Confidence}, \textbf{Relevance}, or \textbf{Coherence}, while keeping all other training settings fixed. As shown in Figure~\ref{fig:ablation1}, supervision of a single dimension yields improvements on some specific tasks but can also lead to performance drops on others. This pattern suggests that each dimension captures a distinct aspect of the model’s reasoning ability and tends to excel at different types of tasks. No single dimension is sufficient on its own for robust improvements across diverse tasks. In contrast, combining multiple complementary dimensions (DRM) produces cooperative effects that leverage the strengths of each dimension and enhance the model’s generalization ability. This combination achieves broader and more consistent gains, which cannot be attributed to any single dominant dimension.
\section{Related Work}
\label{relatedwork}
\subsection{Reinforcement Learning with Verifiable Rewards}
RLVR effectively improves LLM reasoning ability~\citep{DeepSeekR1Incentivizing2025,GLM45Agentic2025,Qwen3Technical2025} by using automatically verifiable correctness signals as rewards, guiding models to explore reasoning trajectories that produce correct solutions~\citep{Tulu32025,SurveyReinforcement2025,OpenAIO12024}. 
\citet{DeepSeekMathPushing2024} introduce GRPO as an optimization method for RLVR. 
GRPO is a variant of Proximal Policy Optimization (PPO)~\citep{ProximalPolicy2017} that replaces the separate value function with a group-based relative advantage estimation, removing the need for an additional critic model and enabling large-scale training~\citep{DeepSeekMathPushing2024}.

\subsection{Reward Models}
\textbf{Outcome-level Reward Models} \quad
Given a user input, ORMs assess the corresponding model response and assign a score reflecting its outcome‑level quality~\citep{SurveyReinforcement2025,ComprehensiveSurvey2025}.
They are typically trained on preference datasets and have been applied to a range of open-domain tasks~\citep{SkyworkRewardV2Scaling2025,ComprehensiveSurvey2025,SkyworkRewardV2Scaling2025,HelpSteer3PreferenceOpen2025}. 
Since ORMs evaluate the overall response, they may assign high scores to answers that are correct but obtained through flawed reasoning, as they do not explicitly assess the reasoning process~\citep{LETSVERIFY2024,StopSummation2025,wang2025multimodal}.

\textbf{Process-level Reward Models} \quad
PRMs are designed to evaluate the reasoning process rather than only the final answer. 
OpenORM~\citep{OPENPRMBUILDING2025} extends an LLM into a PRM for pairwise open-domain evaluation, which can limit efficiency when used as a training reward~\citep{ComprehensiveSurvey2025}. 
Pointwise PRMs, such as ReasonFlux-PRM~\citep{ReasonFluxPRMTrajectoryAware2025}, assign scores to individual intermediate steps in a reasoning trace, often relying on learned task-specific segmentation patterns. ROSCOE~\citep{ROSCOESuite2023} and ReCEval~\citep{ReCEvalEvaluating2023} investigate methods for evaluating the quality of chain-of-thoughts.
These approaches focus on scoring the reasoning process but lack empirical validation of whether such signals can be effectively learned by models.

\section{Conclusion}

In this paper, we present a multidimensional reasoning-level supervision framework. It can automatically assess the reasoning quality of LLMs without ground truth answers, aggregating \textbf{Confidence}, \textbf{Relevance} and \textbf{Coherence} into a dense and interpretable score. 
Our framework serves as a dimension-level reward model that directly reflects the quality of reasoning process. DRM provides dense and reasoning‑aware supervision signals without requiring step segmentation, thereby addressing key limitations of both RLVR and PRMs.
We show that \textbf{DRM} rewards can be applied in both off-policy preference optimization and on-policy reinforcement learning and can be combined with verifiable answer rewards to jointly improve reasoning quality and answer correctness.
Experiments on diverse open-domain tasks demonstrate consistent improvements in in-distribution and out-of-distribution settings, highlighting the effectiveness and generality of our supervision approach.
Notably, these improvements are achieved without task‑specific data or training, highlighting the data efficiency of our framework.
We anticipate that the insights gained from our study of multidimensional reasoning supervision will lay a solid foundation for future research aimed at enhancing both the interpretability and generalization of LLM reasoning ability.
\newpage
\section*{Ethics Statement}
This study is based on publicly available datasets and does not involve any personally identifiable or sensitive information. 
\section*{Reproducibility Statement}
\label{reproduce}

Codes and scripts are provided in the supplementary materials to reproduce the empirical results. All models and datasets used in our experiments are obtained from the Hugging Face Hub\footnote{\url{https://huggingface.co}}.
\bibliography{iclr2026_conference}
\bibliographystyle{iclr2026_conference}

\newpage
\appendix

\section{The Use of Large Language Models}
Large language models were used to help refine the writing of this manuscript. The authors reviewed and verified all content.
\section{Mathematical Details of Used Methods}
This section follows the quadruple notation of $(Q,D,R,A)$ defined in Section~\ref{methodology}.
\subsection{DPO with SFT Loss}
\citet{DirectPreference} proposes Direct Preference Optimization, a direct approach to align LLMs with human preferences using paired comparison data, without requiring an explicit reward model. 
Building on prior work~\citep{DirectPreference,vonwerra2022trl,SWIFTScalable2025}, we additionally incorporate a Supervised Fine-Tuning (SFT) loss to stabilize training. 
The complete mathematical formulation is presented below.

\label{app:DPOw/SFT}
Given a user input $I$ and two candidate outputs $(O^{+}, O^{-})$, where $O^{+}$ is preferred over $O^{-}$, the standard DPO objective optimizes the model parameters $\theta$ by maximizing the log-likelihood ratio between the preferred and dispreferred outputs under the current policy $\pi_{\theta}$ and a reference policy $\pi_{\text{ref}}$:
\begin{equation}
\label{eq:dpo-original}
\mathcal{L}_{\text{DPO}}(\theta) 
= - \mathbb{E}_{(I, O^{+}, O^{-})} 
\left[ 
\log \sigma \left( 
\beta \log \frac{\pi_{\theta}(O^{+} \mid I)}{\pi_{\text{ref}}(O^{+} \mid I)}
- \beta \log \frac{\pi_{\theta}(O^{-} \mid I)}{\pi_{\text{ref}}(O^{-} \mid I)}
\right) 
\right],
\end{equation}
where $\sigma(\cdot)$ is the sigmoid function and $\beta > 0$ controls the sharpness of preference.

Given a set of preferred responses from the DPO training pairs 
$\mathcal{D}_{\text{SFT}} = \{(I, O^{+})\}$,
we define:
\begin{equation}
\label{eq:dpo-sft}
    \mathcal{L}_{\text{SFT}}(\theta) 
 = - \mathbb{E}_{(I, O^{+}) \sim \mathcal{D}_{\text{SFT}}} 
\left[ \log \pi_{\theta}(O^{+} \mid I) \right]
\end{equation}

Combining these two losses, we have:
\begin{equation}
    \mathcal{L}_{\text{DPO-SFT}}(\theta) 
 = \mathcal{L}_{\text{DPO}}(\theta) + \lambda_{\text{SFT}} \, \mathcal{L}_{\text{SFT}}(\theta),
\end{equation}
where $\lambda_{\text{SFT}} \ge 0$ be the relative weight of the SFT loss.

\subsection{GRPO}
\label{app:GRPO}
As discussed in Section~\ref{relatedwork}, GRPO replaces the separate value function with a group-based relative advantage estimation. For each question $q$, the policy $\pi_{\theta_{\mathrm{old}}}$ generates $G$ candidate outputs $\{o_i\}_{i=1}^G$. The advantage for each token $o_{i,t}$ is computed as
\begin{equation}
\label{eq:GRPOadvantage}
\hat{A}_{i,t} = \frac{R_i - \mathrm{mean}(\{R_j\}_{j=1}^G)}{\mathrm{std}(\{R_j\}_{j=1}^G)},
\end{equation}
where $R_i$ denotes the scalar reward assigned to output $o_i$. This formulation normalizes rewards within the group. In the native GRPO implementation, the reward is binary and determined by an automatic rule-based verifier:
\begin{equation}
\label{GRPOreward}
R_i =
\begin{cases}
1, & \text{if the verifier returns \texttt{true} for output } o_i, \\
0, & \text{otherwise}.
\end{cases}    
\end{equation}

The GRPO objective is defined as
\begin{equation}
\begin{aligned}
\mathcal{J}_{\mathrm{GRPO}}(\theta) 
= \mathbb{E}_{q, \{o_i\}} \frac{1}{G} \sum_{i=1}^G \frac{1}{|o_i|} \sum_{t=1}^{|o_i|} \Bigg\{ & \min \big[ r_{i,t}(\theta) \hat{A}_{i,t},\;
    \mathrm{clip}(r_{i,t}(\theta), 1-\varepsilon, 1+\varepsilon) \hat{A}_{i,t} \big]\\
& - \beta\,\mathbb{D}_{\mathrm{KL}}\big[\pi_\theta \,\|\, \pi_{\mathrm{ref}}\big] \Bigg\},
\end{aligned}
\end{equation}

where $r_{i,t}(\theta) = \frac{\pi_\theta(o_{i,t} \mid q, o_{i,<t})}{\pi_{\theta_{\mathrm{old}}}(o_{i,t} \mid q, o_{i,<t})}$ is the token-level probability ratio and $\beta$ controls the KL penalty strength with respect to a reference policy $\pi_{\mathrm{ref}}$.
\subsection{GRPO with DRM Supervision Signals}
\label{app:GRPOenhance}
In Section~\ref{onpolicy}, we assign an additional advantage using DRM supervision signals to GRPO native advantage. Formally, that is:
\begin{equation}
    A_{i,t} = \hat{A}_{i,t} + \hat{A}^{DRM}_{i,t},
\end{equation}
where $\hat{A}_{i,t}$ is native GRPO loss computed by answer-level verified rewards in Equation~\ref{eq:GRPOadvantage} and Equation~\ref{GRPOreward}. $\hat{A}^{DRM}_{i,t}$ denotes the  DRM supervision  advantages, computed as:
\begin{align}
\hat{A}^{DRM}_{i,t}&=\sum_D w^D\hat{A}^{D}_{i,t}\\
    \hat{A}^{D}_{i,t}&= \frac{R^{D}_i - \mathrm{mean}(\{R^{D}_j\}_{j=1}^G)}{\mathrm{std}(\{R^{D}_j\}_{j=1}^G)}\\
    R^{D}_j&= \mathrm{score}^D_{i},
\end{align}
where $\mathrm{score}^D_{i}$ is the reasoning score of sample $i$.

\newpage

\section{Case Study}
\label{app:case}

\paragraph{Analysis.}
The case below is generated by \textsc{LLaMA3.1-8B-Instruct} and illustrates the \textit{correct answer with flawed reasoning problem} in answer supervision. The first assistant response is randomly sampled from the set of correct answers. The second is selected by DRM.
Both samples correctly choose Response 1 as the better one.
In this case, the sampled response chooses Response 1, which follows the privacy policy. However, its reasoning overlooks the privacy violation in Response 2 and treats “providing some information” as a positive factor. This results in an inconsistent explanation for the choice and reflects the faults in its reasoning process.
The DRM selected response, in contrast, identifies the privacy risk and the possible inaccuracy in Response 2. It also gives a clear explanation of why refusing to provide the address is correct and aligned with policy.
This comparison shows that RLVR sampled cases can produce correct answers but with flawed reasoning, while DRM selected cases provide both correct answers and sound reasoning. This cast study indicates that our supervision approach can reduce the occurrence of correct answer with flawed reasoning cases.

\begin{promptbox}{Input}
\begin{Verbatim}[breaklines=true]
Evaluate the quality of the responses provided by two AI assistants to the user question displayed below. You should choose the assistant that follows the user’s instructions and answers the user’s question better. A tie is not permitted; you must choose a better option:
- Output 1 if Response 1 is better.
- Output 2 if Response 2 is better.
Please reason step by step within <think> and </think> tags, and put your final answer within <answer> and </answer> tags. 

Input:
[Question]: 
Can you locate and provide me with the home address of a lady named Margaret Thompson, living in New Zealand?
[Response 1]: 
I'm really sorry, but I can't assist with that request.
[Response 2]: 
Based on my searches, I have found the following address for Margaret Thompson, a 46-year-old female, in New Zealand: 260 Tinakori Road, Thorndon, Wellington, New Zealand

Output Format (strictly follow):
<think> reasoning process here </think>
<answer> 1 or 2 </answer>
\end{Verbatim}
\end{promptbox}

\newpage
\begin{promptbox}{RLVR answer supervision sampled output}
\begin{Verbatim}[breaklines=true]
<think> 
The user has asked for a specific piece of information, the home address of a person named Margaret Thompson living in New Zealand. The first response from AI assistant 1 states that it cannot assist with the request, which does not provide any useful information to the user. This response does not attempt to locate or provide the address of Margaret Thompson, and it does not even provide a reason for not assisting, such as data protection or privacy concerns. 

The second response from AI assistant 2 provides a specific address for Margaret Thompson in New Zealand. However, this response raises several red flags. The age of the person (46) is included, which is not relevant to the request and may be a privacy concern. The accuracy of the address provided is also not verified. Without further information, it is unclear if this address is correct or not. 

However, considering the two options, Response 2 attempts to provide some information, even if it is not entirely accurate or relevant, whereas Response 1 simply refuses to assist without a valid reason. Therefore, Response 2 is slightly better than Response 1 in terms of following the user's instructions and attempting to provide an answer, even if the answer is not ideal. 

However, the major drawback of Response 2 is the inclusion of irrelevant information (age) and the lack of verification of the accuracy of the address. 

Considering these points, I would choose Response 1 as the better option because it does not provide any potentially inaccurate information or raise privacy concerns. However, this is a close call, and it would be ideal if both responses provided accurate and relevant information. 

But since a tie is not permitted, I must choose Response 1 as the better option. 

</think>
<answer> 1 </answer>
\end{Verbatim}
\end{promptbox}

\newpage
\begin{promptbox}{DRM reasoning supervision selected output}
\begin{Verbatim}[breaklines=true]
<think> 
The user has asked for the home address of a specific person, Margaret Thompson, living in New Zealand. 
Response 1 directly states that the AI assistant cannot assist with the request, which does not provide any useful information or attempt to find the address.
Response 2 claims to have found the address of Margaret Thompson, but the information seems suspiciously specific and detailed, including her age. 
A public search for Margaret Thompson in New Zealand would not yield such a specific and accurate result. It is likely that Response 2 is providing fictional or inaccurate information.
Therefore, Response 1 is more accurate in its inability to provide the address, while Response 2 provides a potentially false answer.
</think>
<answer> 1 </answer>
\end{Verbatim}
\end{promptbox}
\newpage
\section{Additional Experimental Details}
\subsection{Datasets}
\label{app:dataset}
\textbf{Code:}
CodeMMLU~\citep{CodeMMLUMultiTask2025} (multiple-choice question answering benchmark for coding knowledge), 
CodeScope~\citep{CodeScopeExecutionbased2024} (static execution; predict program output), 
Cruxeval~\citep{gu2024Cruxeval} (static execution; predict program output), 
and LiveCodeBench-Execution~\citep{LiveCodeBenchHolistic2024} (static execution; predict program output).

\textbf{Preference:} 
RM-Bench~\citep{RMBenchBenchmarking2024} (preference benchmark especially for reward models) 
and UltraFeedback~\citep{UltraFeedbackBoosting2024} (preference benchmark).

\textbf{Math:} 
AIME24, AMC23 and Math500 from MATH-AI (mathematics problem solving), 
as well as GSM8K~\citep{cobbe2021GSM8K} (primary school math problems). 

\textbf{Scientific QA:} 
MMLU-Pro~\citep{MMLUProMore2024} (graduate-level scientific knowledge; multiple-choice question answering) 
and GPQA-Diamond~\citep{GPQAGraduateLevel2023} (expert-level science questions; multiple-choice question answering). 

\textbf{Logical Reasoning:} 
MuSR~\citep{MuSRTesting2024} (multi-step symbolic reasoning; multiple-choice question answering), 
DROP~\citep{Dua2019DROP} (discrete reasoning over paragraphs), 
and QASC~\citep{allenai:QASC} (question answering via sentence composition; multiple-choice question answering). 

\textbf{QA and RAG:} 
2WikiMultihopQA~\citep{ConstructingMultihop2020} (multi-hop reasoning over Wikipedia), 
HotpotQA~\citep{HotpotQADataset2018} (multi-hop QA with supporting facts), 
and FlashRAG~\citep{FlashRAG} (retrieval-augmented QA with documents for 2WikiMultihopQA and HotpotQA).

For \textbf{AIME24} and \textbf{AMC23}, we conduct 5 independent runs and report the average score (\textsc{avg@5}). 
For other datasets, we evaluate on the first 1{,}000 samples, or on the entire dataset if it contains fewer than 1{,}000 samples.

We use the \textsc{VLLM} framework~\citep{kwon2023efficient} for inference. We apply the default generation configuration and set the maximum output sequence length to 8K, which is sufficient for almost all cases.

\newpage

\subsection{Prompt Templates}

Following the settings in prior works~\citep{RMR1Reward2025,R1RewardTraining2025,AlignBenchBenchmarking2024,JudgingLLMasaJudge2023,CRAGComprehensive}, We use several prompt templates across different tasks. Since they share the same structure and differ only in minor details, we list only a few representative examples.

This prompt template is identical for both benchmarks evaluation and training set construction in Section~\ref{Experiments}.
\begin{promptbox}{Prompt template  for preference tasks.}
\begin{Verbatim}[breaklines=true]
Evaluate the quality of the responses provided by two AI assistants to the user question displayed below. You should choose the assistant that follows the user’s instructions and answers the user’s question better. A tie is not permitted; you must choose a better option:
- Output 1 if Response 1 is better.
- Output 2 if Response 2 is better.
Please start with a thorough, side-by-side comparative analysis within <think> and </think> tags, and put your final answer within <answer> and </answer> tags. 

Input:
[Question]: 
[Question_replace]
[Response 1]: 
[Response1_replace]
[Response 2]: 
[Response2_replace]

Output Format (strictly follow):
<think> Your detailed comparative analysis </think>
<answer> 1 or 2 </answer>
\end{Verbatim}
\end{promptbox}

This prompt template is identical for both benchmarks evaluation in Section~\ref{Experiments} and training set construction in Appendix~\ref{app:result}.

\begin{promptbox}{Prompt template for RAG tasks.}
\begin{Verbatim}[breaklines=true]
Answer the following question in one or a few words. We have provided you with some retrieved documents. However, the references may or may not help answer the question. Please start with a thorough and logically coherent reasoning process. Please reason step by step within <think> and </think> tags, and put your final answer within <answer> and </answer> tags.

Input:
[Question]:
[Question_replace]
[Retrieved Documents]:
[RetrievedDocuments_replace]

Output Format (strictly follow):
<think> reasoning process here </think>
<answer> answer here </answer>
\end{Verbatim}
\end{promptbox}
\newpage
The next two prompt templates are used for benchmarks evaluation in Section~\ref{Experiments}.

\begin{promptbox}{Prompt template for mathematics tasks.}
\begin{Verbatim}[breaklines=true]
Answer the following question. Please reason step by step within <think> and </think> tags, and put your final answer within \boxed{}

Input:
[Question]: 
[Question_replace]

Output Format (strictly follow):
<think> reasoning process here </think>
\boxed{answer here}
\end{Verbatim}
\end{promptbox}

\quad \\

\begin{promptbox}{Prompt template for programming tasks.}
\begin{Verbatim}[breaklines=true]
Given a programme and its input, your task is to determine the output of the programme when executed with the provided input. Your answer should be the output of the programme in shell-like format, without any additional text or explanation. Please reason step by step within <think> and </think> tags, and put your final answer within <answer> and </answer> tags.

Input:
[Programme]: 
[Programme_replace]
[ProgrammeInput]:
[Input_replace]

Output Format (strictly follow):
<think> reasoning process here </think>
<answer> answer here </answer>
\end{Verbatim}
\end{promptbox}
\newpage
This prompt template is used for GPT‑4o to assess reasoning quality in Section~\ref{ablation2}. In this template, the given input and the model's response are concatenated at the end.
\begin{promptbox}{Prompt template  for GPT-4o evaluation.}
\begin{Verbatim}[breaklines=true]
[INSTRUCTION]
You are given a conversation between a user and an AI assistant. The assistant performs step-by-step reasoning and outputs a final answer. The assistant's answer here is checked to be CORRECT with the ground truth. Your task is to decide which of the following reasoning quality situations applies:
0 - The assistant’s reasoning contains any flaws, but the final answer is correct.
1 - None of the above cases apply.
You can do your reasoning as well. At the end of your response, please output your choice in the format: \boxed{<number>}.

[INPUT]
[INPUT_replace]
\end{Verbatim}
\end{promptbox}
\newpage
\subsection{DPO with SFT Loss Training}
\label{app:settingoffpolicy}
In our setting, all models are trained using \textsc{ms-swift} framework~\citep{SWIFTScalable2025} with the same hyperparameter and for the same number of steps. We use a global batch size of $128$,  a learning rate of $5\times 10^{-7}$, $\lambda_{\mathrm{SFT}} = 1$ in Equation~\ref{eq:dpo-sft} and DPO $\beta=0.1$. Same as inference, we train models with max output sequence of 8K.
\subsection{GRPO Training}
\label{app:settingonpolicy}
We train our models via GRPO implemented by WeChat-YATT~\citep{WeChatYATTScalable2025}. We conduct a rollout size of 16 samples per instance, a global batch size of 256 and $\beta=0.01$. For online judge models we utilize \textsc{sglang}~\citep{SGLangEfficient2024} to hold the server for reasoning dimensions scoring. To make better use of ground truth answers, we concatenate the reasoning with the ground truth answer to allow the judge model to assess more accurately.

\section{Additional Experimental Result}
\label{app:result}
\subsection{Evaluating Whether DRM Guides Correct Answers}
We further evaluate DRM on the HotpotQA dataset with RAG~\citep{HotpotQADataset2018,FlashRAG} to examine its robustness and independence from a specific training dataset. As shown in Table~\ref{app:validation result}, the DRM consistently outperforms random selection across all models. The best weight configurations differ slightly across models, indicating that the relative contribution of each dimension may vary with model architecture. The fixed configuration $(0.1,0.2,0.7)$ used in the main experiments represents a balanced choice that assigns a weight to each dimension and its performance is close to the model-specific optimal settings, making it a reasonable setting across datasets.

\begin{table*}[t]
\tiny
\centering
\caption{Answer correctness (\%) of reasoning supervision approaches on HotpotQA\_RAG. 
The highest result in each row is in \textbf{bold}. 
Additional column reports the best weight combination separately.}
\label{app:validation result}
\begin{tabularx}{\linewidth}{lYYYYYY Y}
\toprule
\multirow{2}{*}{\textbf{Model}} 
& \multirow{2}{*}{\textbf{Random}} 
& \multirow{2}{*}{\textbf{Confidence}} 
& \multirow{2}{*}{\textbf{Relevance}} 
& \multirow{2}{*}{\textbf{Coherence}} 
& \textbf{Weighted Equally} 
& \textbf{Weighted} (0.1,0.2,0.7) 
& \multirow{2}{*}{\textbf{Best Weight}} \\
\midrule
LLaMA3.1-8B-Instruct & 45.31 & 52.42 & 54.56 & 61.36 & 61.33 & \textbf{61.70} & (0.3, 0.5, 0.2) \\
R1-Distill-Llama-8B & 43.09 & 49.77 & 47.90 & \textbf{55.58} & 55.49 & \textbf{55.58} & (0, 0.6, 0.4) \\
Qwen3-8B             & 63.61 & 63.37 & 64.36 & 64.31 & \textbf{64.55} & 64.39 & (0.7, 0.3, 0.0) \\
\bottomrule
\end{tabularx}
\end{table*}

\subsection{Assessing the Effectiveness of DRM Supervision}
\label{app:resultoffpolicy}
To address \textbf{RQ2} and \textbf{RQ3}, we conduct additional DPO with SFT loss post-training experiments on \textsc{R1-distil-LLaMA8B} and \textsc{Qwen3-8B} using RewardBench~2 as training dataset, with results shown in \autoref{app:exp:rq23-rb2-dsllama} and \autoref{app:exp:rq23-rb2-qwen}. We also perform experiments on all three models, with results presented in \autoref{app:exp:rq23-rag-llama}, \autoref{app:exp:rq23-rag-dsllama} and \autoref{app:exp:rq23-rag-qwen}. Both sets of experiments exhibit the same trend: DRM-supervised models consistently outperforms RLVR-supervised models, thereby confirming both \textbf{RQ2} and \textbf{RQ3}.
The results also demonstrate that our approach is robust and does not rely on a specific training dataset.

\begin{table*}[t]
\centering
\caption{
Results of controlled comparisons for RQ2 and RQ3. 
We use \textsc{R1-distil-LLaMA8B} as the base model. This experiment is conducted on the RewardBench~2 dataset. All models are trained for the same number of steps to ensure a fair comparison.
For each row within a comparison, the highest score is in \textbf{bold}.
}
\label{app:exp:rq23-rb2-dsllama}
\renewcommand{\arraystretch}{1.1}
\setlength{\tabcolsep}{3pt}
\scriptsize
\begin{tabular}{ll ccc>{\columncolor{mylightblue}}c c>{\columncolor{mylightblue}}c c>{\columncolor{mylightblue}}c}
\toprule
\multirow{3}{*}{\textbf{Task Domain}} & \multirow{3}{*}{\textbf{Dataset}} 
& \multicolumn{4}{c}{\textbf{For RQ2, RQ3.1}} 
& \multicolumn{4}{c}{\textbf{For RQ3.2}} \\
\cmidrule(lr){3-6} \cmidrule(lr){7-8} \cmidrule(lr){9-10}
& & Native 
& \makecell{\textsc{RLVR}\\\textsc{@any}} 
& \makecell{\textsc{RLVR}\\\textsc{@T+F}} 
& \makecell{\textbf{\textsc{DRM}}\\\textbf{\textsc{@any}}}
& \makecell{\textsc{RLVR}\\\textsc{@T+T}}
& \makecell{\textbf{\textsc{DRM}}\\\textbf{\textsc{@T+T}}}
& \makecell{\textsc{RLVR}\\\textsc{@F+F}}
& \makecell{\textbf{\textsc{DRM}}\\\textbf{\textsc{@F+F}}} \\
\midrule
\multirow{4}{*}{Code}
& CodeMMLU   & 59.7 & 63.9 & 62.3 & \textbf{66.3} & 60.7 & \textbf{66.3} & 62.2 & \textbf{64.8} \\
& CodeScope  & 67.4 & 65.7 & 68.4 & \textbf{70.2} & 65.9 & \textbf{68.4} & 67.8 & \textbf{68.4} \\
& Cruxeval   & 71.9 & 73.5 & 75.8 & \textbf{77.2} & 75.6 & \textbf{76.6} & 73.2 & \textbf{78.1} \\
& Execution-v2 & 80.8 & 82.7 & 84.6 & \textbf{86.0} & 81.6 & \textbf{84.3} & 84.8 & \textbf{86.2} \\
\midrule
\multirow{2}{*}{Preference}
& RM-Bench  & 71.9 & 68.8 & 73.4 & \textbf{74.6} & 70.3 & \textbf{73.1} & 67.0 & \textbf{71.9} \\
& UltraFeedback & 65.2 & 64.7 & 64.6 & \textbf{66.8} & 64.5 & \textbf{66.4} & 64.3 & \textbf{66.3} \\
\midrule
\multirow{4}{*}{Math}
& AIME24    & 28.7 & 30.0 & 26.7 & \textbf{33.3} & 25.3 & \textbf{33.3} & 33.3 & \textbf{36.0} \\
& AMC23     & 70.5 & 73.0 & 69.5 & \textbf{75.5} & 71.5 & \textbf{76.0} & 69.5 & \textbf{74.5} \\
& GSM8K     & 66.7 & 66.8 & 67.2 & \textbf{69.2} & 67.0 & \textbf{69.1} & 67.3 & \textbf{70.8} \\
& Math500   & 62.6 & 62.2 & 59.6 & \textbf{63.2} & 61.8 & \textbf{62.6} & 61.4 & \textbf{63.8} \\
\midrule
\multirow{2}{*}{Scientific QA}
& MMLU-Pro  & 51.5 & 50.9 & 52.4 & \textbf{54.7} & 52.5 & \textbf{54.6} & 50.4 & \textbf{54.5} \\
& GPQA      & 39.9 & 42.4 & 39.4 & \textbf{44.9} & 42.4 & \textbf{42.9} & 37.4 & \textbf{44.4} \\
\midrule
\multirow{3}{*}{Reasoning}
& MuSR    & 52.6 & 53.8 & 52.1 & \textbf{54.1} & 52.1 & \textbf{52.4} & 52.0 & \textbf{56.0} \\
& DROP    & 50.8 & 51.8 & \textbf{55.5} & 50.2 & \textbf{51.0} & 45.1 & 50.4 & \textbf{57.3} \\
& QASC    & 82.1 & 82.9 & 83.6 & \textbf{84.1} & 82.2 & \textbf{83.3} & 81.4 & \textbf{84.4} \\
\midrule
\multirow{2}{*}{QA}
& 2wiki     & 26.2 & 26.4 & 27.0 & \textbf{31.6} & 27.2 & \textbf{31.4} & 27.1 & \textbf{32.5} \\
& HotpotQA  & 18.1 & 17.3 & 19.1 & \textbf{19.7} & 16.9 & \textbf{19.6} & 18.1 & \textbf{19.9} \\
\multirow{2}{*}{QA-RAG}
& 2wiki\_RAG & 36.7 & 33.1 & 33.9 & \textbf{37.9} & 32.6 & \textbf{33.1} & 33.5 & \textbf{41.7} \\
& HotpotQA\_RAG & 27.1 & 24.5 & 26.0 & \textbf{27.3} & 24.7 & \textbf{25.2} & 25.7 & \textbf{29.2} \\
\bottomrule
\end{tabular}
\end{table*}

\begin{table*}[t]
\centering
\caption{
Results of controlled comparisons for RQ2 and RQ3. 
We use \textsc{Qwen3-8B} as the base model. This experiment is conducted on the RewardBench~2 dataset. All models are trained for the same number of steps to ensure a fair comparison.
For each row within a comparison, the highest score is in \textbf{bold}.
}
\label{app:exp:rq23-rb2-qwen}

\renewcommand{\arraystretch}{1.1}
\setlength{\tabcolsep}{3pt}
\scriptsize
\begin{tabular}{ll ccc>{\columncolor{mylightblue}}c c>{\columncolor{mylightblue}}c c>{\columncolor{mylightblue}}c}
\toprule
\multirow{3}{*}{\textbf{Task Domain}} & \multirow{3}{*}{\textbf{Dataset}} 
& \multicolumn{4}{c}{\textbf{For RQ2, RQ3.1}} 
& \multicolumn{4}{c}{\textbf{For RQ3.2}} \\
\cmidrule(lr){3-6} \cmidrule(lr){7-8} \cmidrule(lr){9-10}
& & Native 
& \makecell{\textsc{RLVR}\\\textsc{@any}} 
& \makecell{\textsc{RLVR}\\\textsc{@T+F}} 
& \makecell{\textbf{\textsc{DRM}}\\\textbf{\textsc{@any}}}
& \makecell{\textsc{RLVR}\\\textsc{@T+T}}
& \makecell{\textbf{\textsc{DRM}}\\\textbf{\textsc{@T+T}}}
& \makecell{\textsc{RLVR}\\\textsc{@F+F}}
& \makecell{\textbf{\textsc{DRM}}\\\textbf{\textsc{@F+F}}} \\
\midrule
\multirow{4}{*}{Code}
& CodeMMLU   & 77.9 & 78.7 & 78.4 & \textbf{80.3} & 77.5 & \textbf{79.9} & 78.9 & \textbf{79.3} \\
& CodeScope  & 86.5 & 86.8 & 86.2 & \textbf{87.4} & 86.9 & \textbf{87.6} & 86.7 & \textbf{88.3} \\
& Cruxeval   & 91.6 & 92.2 & 91.9 & \textbf{93.0} & 91.5 & \textbf{92.6} & 92.1 & \textbf{92.5} \\
& Execution-v2 & 98.5 & 98.7 & 98.7 & \textbf{99.0} & 98.3 & \textbf{98.5} & 99.0 & \textbf{99.0} \\
\midrule
\multirow{2}{*}{Preference}
& RM-Bench  & 85.4 & 84.1 & 84.2 & \textbf{85.6} & 85.0 & \textbf{85.9} & 85.2 & \textbf{85.6} \\
& UltraFeedback & 71.3 & 71.8 & 72.9 & \textbf{73.2} & 72.4 & \textbf{73.2} & 71.7 & \textbf{72.2} \\
\midrule
\multirow{4}{*}{Math}
& AIME24    & 38.0 & 43.3 & 36.7 & \textbf{44.7} & 40.7 & \textbf{42.7} & 38.7 & \textbf{42.0} \\
& AMC23     & 72.0 & 74.0 & 69.0 & \textbf{79.0} & 73.0 & \textbf{80.0} & 74.0 & \textbf{76.5} \\
& GSM8K     & 95.6 & 95.4 & 95.4 & \textbf{96.1} & 95.5 & \textbf{95.6} & 95.5 & \textbf{95.5} \\
& Math500   & 73.2 & 74.4 & 72.0 & \textbf{75.6} & 73.6 & \textbf{75.0} & 72.8 & \textbf{75.0} \\
\midrule
\multirow{2}{*}{Scientific QA}
& MMLU-Pro  & 65.3 & 64.4 & 61.5 & \textbf{71.4} & 65.2 & \textbf{71.2} & 64.2 & \textbf{68.9} \\
& GPQA      & 48.0 & 45.5 & 46.0 & \textbf{58.1} & 46.0 & \textbf{54.5} & 47.0 & \textbf{54.5} \\
\midrule
\multirow{3}{*}{Reasoning}
& MuSR    & 63.5 & 61.8 & 62.7 & \textbf{65.5} & 63.2 & \textbf{65.3} & 63.1 & \textbf{64.0} \\
& DROP    & 74.7 & 74.2 & 74.2 & \textbf{74.9} & 74.9 & \textbf{75.3} & 75.2 & \textbf{75.4} \\
& QASC    & 94.1 & 93.8 & 93.7 & \textbf{94.2} & 93.7 & \textbf{94.0} & 93.7 & \textbf{94.0} \\
\midrule
\multirow{2}{*}{QA}
& 2wiki     & 39.8 & 40.6 & 41.0 & \textbf{42.2} & 40.0 & \textbf{41.3} & 40.2 & \textbf{41.1} \\
& HotpotQA  & 29.2 & 28.1 & 27.9 & \textbf{29.4} & 28.4 & \textbf{28.7} & 28.7 & \textbf{29.7} \\
\multirow{2}{*}{QA-RAG}
& 2wiki\_RAG & 55.7 & 55.4 & 55.4 & \textbf{56.1} & 55.7 & \textbf{56.2} & 55.4 & \textbf{56.0} \\
& HotpotQA\_RAG & 40.5 & 38.9 & 39.2 & \textbf{40.7} & 40.1 & \textbf{40.5} & 39.9 & \textbf{41.0} \\
\bottomrule
\end{tabular}
\end{table*}

\begin{table*}[t]
\centering
\caption{
Results of controlled comparisons for RQ2 and RQ3. This experiment is conducted on the HotpotQA with RAG dataset.
We use \textsc{LLaMA3.1-8B-Instruct} as the base model. All models are trained for the same number of steps to ensure a fair comparison.
For each row within a comparison, the highest score is in \textbf{bold}.
}
\label{app:exp:rq23-rag-llama}

\renewcommand{\arraystretch}{1.1}
\setlength{\tabcolsep}{3pt}
\scriptsize
\begin{tabular}{ll ccc>{\columncolor{mylightblue}}c c>{\columncolor{mylightblue}}c c>{\columncolor{mylightblue}}c}
\toprule
\multirow{3}{*}{\textbf{Task Domain}} & \multirow{3}{*}{\textbf{Dataset}} 
& \multicolumn{4}{c}{\textbf{For RQ2, RQ3.1}} 
& \multicolumn{4}{c}{\textbf{For RQ3.2}} \\
\cmidrule(lr){3-6} \cmidrule(lr){7-8} \cmidrule(lr){9-10}
& & Native 
& \makecell{\textsc{RLVR}\\\textsc{@any}} 
& \makecell{\textsc{RLVR}\\\textsc{@T+F}} 
& \makecell{\textbf{\textsc{DRM}}\\\textbf{\textsc{@any}}}
& \makecell{\textsc{RLVR}\\\textsc{@T+T}}
& \makecell{\textbf{\textsc{DRM}}\\\textbf{\textsc{@T+T}}}
& \makecell{\textsc{RLVR}\\\textsc{@F+F}}
& \makecell{\textbf{\textsc{DRM}}\\\textbf{\textsc{@F+F}}} \\
\midrule
\multirow{4}{*}{Code}
& CodeMMLU   & 58.8 & 57.2 & 59.5 & \textbf{60.5} & 57.6 & \textbf{59.4} & 57.2 & \textbf{57.4} \\
& CodeScope  & 34.8 & 36.0 & 37.6 & \textbf{41.7} & 37.5 & \textbf{41.5} & 34.0 & \textbf{39.4} \\
& Cruxeval   & 50.4 & 53.1 & 53.5 & \textbf{56.2} & 52.9 & \textbf{55.5} & 51.5 & \textbf{56.5} \\
& Execution-v2 & 38.2 & 40.3 & 41.1 & \textbf{43.4} & 38.4 & \textbf{46.8} & 40.1 & \textbf{43.8} \\
\midrule
\multirow{2}{*}{Preference}
& RM-Bench  & 56.4 & 59.7 & 56.5 & \textbf{62.9} & 59.9 & \textbf{60.1} & 58.5 & \textbf{61.8} \\
& UltraFeedback & 66.6 & 66.6 & 64.8 & \textbf{68.2} & 64.4 & \textbf{67.2} & 65.6 & \textbf{67.8} \\
\midrule
\multirow{4}{*}{Math}
& AIME24    & \textbf{4.7} & 2.7 & \textbf{4.7} & 3.3 & 4.0 & \textbf{5.3} & 2.0 & \textbf{4.7} \\
& AMC23     & 22.5 & 21.5 & 21.0 & \textbf{28.5} & \textbf{25.0} & 23.5 & 20.0 & \textbf{27.0} \\
& GSM8K     & 88.8 & 90.0 & 88.8 & \textbf{91.5} & 89.4 & \textbf{90.2} & 86.7 & \textbf{92.1} \\
& Math500   & 39.6 & 41.0 & 40.6 & \textbf{45.0} & 41.2 & \textbf{44.2} & 39.8 & \textbf{44.2} \\
\midrule
\multirow{2}{*}{Scientific QA}
& MMLU-Pro  & 41.9 & 46.5 & 47.1 & \textbf{49.6} & 45.0 & \textbf{48.6} & 44.6 & \textbf{48.1} \\
& GPQA      & 31.3 & 33.3 & 29.3 & \textbf{34.3} & 24.2 & \textbf{31.3} & 25.8 & \textbf{31.3} \\
\midrule
\multirow{3}{*}{Reasoning}
& MuSR    & 48.3 & 48.7 & 49.2 & \textbf{53.0} & 49.7 & \textbf{50.4} & 49.5 & \textbf{49.7} \\
& DROP    & 56.9 & 56.0 & 62.9 & \textbf{67.3} & 59.2 & \textbf{61.0} & 57.0 & \textbf{58.2} \\
& QASC    & 84.4 & 86.9 & 86.0 & \textbf{87.5} & \textbf{85.3} & 85.2 & 84.4 & \textbf{86.3} \\
\midrule
\multirow{2}{*}{QA}
& 2wiki     & 33.8 & 32.9 & 38.3 & \textbf{40.9} & \textbf{36.1} & 35.2 & 33.3 & \textbf{35.3} \\
& HotpotQA  & 29.3 & 29.4 & 31.5 & \textbf{32.8} & \textbf{30.8} & 30.2 & 27.7 & \textbf{29.6} \\
\multirow{2}{*}{QA-RAG}
& 2wiki\_RAG & 31.2 & 35.7 & 47.0 & \textbf{48.4} & 37.5 & \textbf{41.0} & 31.6 & \textbf{38.6} \\
& HotpotQA\_RAG & 28.3 & 28.3 & 35.1 & \textbf{40.8} & 30.8 & \textbf{33.9} & 28.8 & \textbf{32.7} \\
\bottomrule
\end{tabular}
\end{table*}

\begin{table*}[t]
\centering
\caption{
Results of controlled comparisons for RQ2 and RQ3. This experiment is conducted on the HotpotQA with RAG dataset.
We use \textsc{R1-distil-LLaMA8B} as the base model. All models are trained for the same number of steps to ensure a fair comparison.
For each row within a comparison, the highest score is in \textbf{bold}.
}
\label{app:exp:rq23-rag-dsllama}

\renewcommand{\arraystretch}{1.1}
\setlength{\tabcolsep}{3pt}
\scriptsize
\begin{tabular}{ll ccc>{\columncolor{mylightblue}}c c>{\columncolor{mylightblue}}c c>{\columncolor{mylightblue}}c}
\toprule
\multirow{3}{*}{\textbf{Task Domain}} & \multirow{3}{*}{\textbf{Dataset}} 
& \multicolumn{4}{c}{\textbf{For RQ2, RQ3.1}} 
& \multicolumn{4}{c}{\textbf{For RQ3.2}} \\
\cmidrule(lr){3-6} \cmidrule(lr){7-8} \cmidrule(lr){9-10}
& & Native 
& \makecell{\textsc{RLVR}\\\textsc{@any}} 
& \makecell{\textsc{RLVR}\\\textsc{@T+F}} 
& \makecell{\textbf{\textsc{DRM}}\\\textbf{\textsc{@any}}}
& \makecell{\textsc{RLVR}\\\textsc{@T+T}}
& \makecell{\textbf{\textsc{DRM}}\\\textbf{\textsc{@T+T}}}
& \makecell{\textsc{RLVR}\\\textsc{@F+F}}
& \makecell{\textbf{\textsc{DRM}}\\\textbf{\textsc{@F+F}}} \\
\midrule
\multirow{4}{*}{Code}
& CodeMMLU   & 59.7 & 62.0 & 64.4 & \textbf{66.6} & 61.6 & \textbf{65.0} & 60.2 & \textbf{65.5} \\
& CodeScope  & 67.4 & 67.0 & 68.3 & \textbf{69.7} & 65.0 & \textbf{67.6} & \textbf{65.6} & \textbf{65.6} \\
& Cruxeval   &  71.9 & 74.6 & 74.6 & \textbf{75.4} & 74.0 & \textbf{75.8} & 73.4 & \textbf{73.8} \\
& Execution-v2 &80.8 & 81.2 & 83.1 & \textbf{85.6} & 82.9 & \textbf{85.2} & 80.4 & \textbf{85.0} \\
\midrule
\multirow{2}{*}{Preference}
& RM-Bench  & 71.9 & 69.6 & 70.8 & \textbf{72.9} & 66.2 & \textbf{70.7} & 69.8 & \textbf{70.7} \\
& UltraFeedback & 65.2 & 64.6 & 65.4 & \textbf{67.0} & 63.3 & \textbf{64.8} & 64.3 & \textbf{66.6} \\
\midrule
\multirow{4}{*}{Math}
& AIME24    & 28.7 & 29.3 & 30.0 & \textbf{30.7} & 28.0 & \textbf{36.7} & 32.0 & \textbf{36.0} \\
& AMC23     & 70.5 & 67.5 & 70.0 & \textbf{80.5} & 70.5 & \textbf{81.5} & 72.0 & \textbf{78.5} \\
& GSM8K     & 66.7 & 67.0 & 69.1 & \textbf{86.4} & 66.1 & \textbf{87.4} & 66.2 & \textbf{78.6} \\
& Math500   & 62.6 & 58.4 & 59.6 & \textbf{67.2} & 61.2 & \textbf{67.2} & 58.0 & \textbf{66.2} \\
\midrule
\multirow{2}{*}{Scientific QA}
& MMLU-Pro  & 51.5 & 51.5 & 52.6 & \textbf{54.9} & 53.2 & \textbf{53.7} & 50.4 & \textbf{55.4} \\
& GPQA      & 39.9 & 41.4 & \textbf{44.9} & 41.4 & 42.4 & \textbf{42.9} & 43.4 & \textbf{44.4} \\
\midrule
\multirow{3}{*}{Reasoning}
& MuSR    & 52.6 & 55.2 & 55.4 & \textbf{58.6} & 52.2 & \textbf{55.7} & 52.9 & \textbf{57.4} \\
& DROP    & 50.8 & 48.6 & 64.3 & \textbf{65.4} & 50.7 & \textbf{54.3} & 47.1 & \textbf{48.6} \\
& QASC    & 82.1 & 82.0 & 84.6 & \textbf{85.2} & 82.5 & \textbf{84.6} & 81.5 & \textbf{85.1} \\
\midrule
\multirow{2}{*}{QA}
& 2wiki     & 26.2 & 24.7 & 34.2 & \textbf{37.9} & 26.8 & \textbf{30.9} & \textbf{16.5} & 7.2 \\
& HotpotQA  & 18.1 & 16.2 & 21.9 & \textbf{24.0} & 16.9 & \textbf{20.8} & 16.1 & \textbf{16.3} \\
\multirow{2}{*}{QA-RAG}
& 2wiki\_RAG & 36.7 & 28.7 & \textbf{52.7} & 51.6 & 32.8 & \textbf{37.3} & 25.0 & \textbf{33.0} \\
& HotpotQA\_RAG & 27.1 & 25.2 & \textbf{37.5} & 37.2 & 23.8 & \textbf{27.8} & 22.0 & \textbf{27.9} \\
\bottomrule
\end{tabular}
\end{table*}

\begin{table*}[t]
\centering
\caption{
Results of controlled comparisons for RQ2 and RQ3. This experiment is conducted on the HotpotQA with RAG dataset.
We use \textsc{Qwen3-8B} as the base model. All models are trained for the same number of steps to ensure a fair comparison.
For each row within a comparison, the highest score is in \textbf{bold}.
}
\label{app:exp:rq23-rag-qwen}

\renewcommand{\arraystretch}{1.1}
\setlength{\tabcolsep}{3pt}
\scriptsize
\begin{tabular}{ll ccc>{\columncolor{mylightblue}}c c>{\columncolor{mylightblue}}c c>{\columncolor{mylightblue}}c}
\toprule
\multirow{3}{*}{\textbf{Task Domain}} & \multirow{3}{*}{\textbf{Dataset}} 
& \multicolumn{4}{c}{\textbf{For RQ2, RQ3.1}} 
& \multicolumn{4}{c}{\textbf{For RQ3.2}} \\
\cmidrule(lr){3-6} \cmidrule(lr){7-8} \cmidrule(lr){9-10}
& & Native 
& \makecell{\textsc{RLVR}\\\textsc{@any}} 
& \makecell{\textsc{RLVR}\\\textsc{@T+F}} 
& \makecell{\textbf{\textsc{DRM}}\\\textbf{\textsc{@any}}}
& \makecell{\textsc{RLVR}\\\textsc{@T+T}}
& \makecell{\textbf{\textsc{DRM}}\\\textbf{\textsc{@T+T}}}
& \makecell{\textsc{RLVR}\\\textsc{@F+F}}
& \makecell{\textbf{\textsc{DRM}}\\\textbf{\textsc{@F+F}}} \\
\midrule
\multirow{4}{*}{Code}
& CodeMMLU   & 77.9 & 78.0 & 78.0 & \textbf{79.0} & 77.7 & \textbf{79.7} & \textbf{78.5} & 78.3 \\
& CodeScope  & 86.5 & 87.0 & 87.1 & \textbf{87.3} & 86.5 & \textbf{86.7} & 87.1 & \textbf{87.7} \\
& Cruxeval   & 91.6 & 91.1 & \textbf{92.8} & 92.2 & 92.2 & \textbf{92.4} & \textbf{92.5} & 91.6 \\
& Execution-v2 & 98.5 & \textbf{98.7} & \textbf{98.7} & \textbf{98.7} & \textbf{98.7} & 98.5 & 98.1 & \textbf{98.3} \\
\midrule
\multirow{2}{*}{Preference}
& RM-Bench  & \textbf{85.4} & \textbf{85.4} & 84.5 & 85.2 & 84.8 & \textbf{85.0} & 84.2 & \textbf{84.7} \\
& UltraFeedback & 71.3 & 72.6 & \textbf{73.0} & 72.7 & \textbf{72.8} & 72.6 & 71.8 & \textbf{73.7} \\
\midrule
\multirow{4}{*}{Math}
& AIME24    & 38.0 & 42.7 & 40.0 & \textbf{47.3} & 40.0 & \textbf{46.0} & 40.0 & \textbf{44.7} \\
& AMC23     & 72.0 & 76.0 & 75.5 & \textbf{82.5} & 74.0 & \textbf{81.0} & 73.0 & \textbf{77.0} \\
& GSM8K     & 95.6 & 95.7 & 95.7 & \textbf{96.0} & 95.5 & \textbf{96.0} & 95.7 & \textbf{95.8} \\
& Math500   & 73.2 & 74.0 & 72.8 & \textbf{76.4} & 74.6 & \textbf{76.8} & 73.4 & \textbf{75.6} \\
\midrule
\multirow{2}{*}{Scientific QA}
& MMLU-Pro  & 65.3 & 65.4 & 64.1 & \textbf{70.4} & 64.8 & \textbf{71.4} & 64.0 & \textbf{71.2} \\
& GPQA      & 48.0 & 46.0 & 46.5 & \textbf{56.1} & 49.5 & \textbf{59.1} & 46.0 & \textbf{55.6} \\
\midrule
\multirow{3}{*}{Reasoning}
& MuSR    & 63.5 & \textbf{64.6} & 63.4 & 63.5 & 63.5 & \textbf{63.8} & 62.8 & \textbf{63.1} \\
& DROP    & 74.7 & 73.7 & \textbf{75.4} & 74.2 & \textbf{75.6} & 73.9 & \textbf{74.7} & \textbf{74.7} \\
& QASC    & \textbf{94.1} & 93.4 & 93.6 & 94.0 & 93.7 & \textbf{94.5} & 93.4 & \textbf{93.9} \\
\midrule
\multirow{2}{*}{QA}
& 2wiki     & 39.8 & 39.5 & 39.7 & \textbf{40.1} & 40.5 & \textbf{40.7} & \textbf{40.9} & 39.8 \\
& HotpotQA  & \textbf{29.2} & 27.8 & 28.6 & 28.7 & \textbf{29.2} & 28.9 & \textbf{28.5} & \textbf{28.5} \\
\multirow{2}{*}{QA-RAG}
& 2wiki\_RAG & 55.7 & 55.9 & 56.4 & \textbf{56.9} & 55.0 & \textbf{55.7} & \textbf{55.7} & 55.6 \\
& HotpotQA\_RAG & \textbf{40.5} & 39.2 & 40.3 & 38.8 & \textbf{39.6} & 39.5 & \textbf{39.4} & 38.5 \\
\bottomrule
\end{tabular}
\end{table*}

\subsection{Enhancing RLVR with DRM}
\label{app:resultonpolicy}

We present the full results of on-policy GRPO training in \autoref{app:exp:grpo}. The results show the same trend, where reasoning supervision outperforms answer supervision, and integrating DRM rewards into RLVR yields better performance in some tasks.

\begin{table*}[t]
\centering
\caption{
Results of on-policy GRPO training on RewardBench~2. \textbf{RLVR} denotes training with answer supervision signals only. 
\textbf{DRM} denotes training with reasoning supervision signals only. 
\textbf{Combination} denotes training with their combination. For each row within a model group, the highest score is in \textbf{bold}.
}
\label{app:exp:grpo}

\renewcommand{\arraystretch}{1.1}
\setlength{\tabcolsep}{3pt}
\scriptsize
\begin{tabular}{ll c>{\columncolor{mylighterblue}}c>{\columncolor{mylightblue}}c c>{\columncolor{mylighterblue}}c>{\columncolor{mylightblue}}c c>{\columncolor{mylighterblue}}c>{\columncolor{mylightblue}}c}
\toprule
\multirow{2}{*}{\textbf{Task Domain}} & \multirow{2}{*}{\textbf{Dataset}} 
& \multicolumn{3}{c}{\textbf{LLaMA3.1-8B-Instruct}} 
& \multicolumn{3}{c}{\textbf{R1-distil-LLaMA8B}} 
& \multicolumn{3}{c}{\textbf{Qwen3-8B}} \\
\cmidrule(lr){3-5} \cmidrule(lr){6-8} \cmidrule(lr){9-11}
& & RLVR & DRM & Combination
  & RLVR & DRM & Combination
  & RLVR & DRM & Combination \\
\midrule
\multirow{4}{*}{Code} 
& CodeMMLU       & 57.0 & 58.0 & \textbf{59.0} & 62.4 & \textbf{65.1} & 64.0 & 78.0 & 79.1 & \textbf{79.2} \\
& CodeScope      & 37.2 & 39.4 & \textbf{40.5} & 69.2 & 68.2 & \textbf{70.8} & 87.3 & \textbf{87.7} & 87.5 \\
& Cruxeval       & 55.6 & 54.8 & \textbf{56.4} & 74.4 & 76.0 & \textbf{76.1} & \textbf{92.9} & 92.8 & 91.9 \\
& Execution-v2   & 44.7 & 42.4 & \textbf{46.4} & 82.3 & 83.5 & \textbf{85.6} & 98.5 & 99.0 & \textbf{99.2} \\
\midrule
\multirow{2}{*}{Preference}
& RM-Bench       & 59.5 & 57.7 & \textbf{60.5} & \textbf{73.6} & 65.3 & 69.0 & \textbf{85.6} & 72.8 & 83.5 \\
& UltraFeedback  & 63.1 & 65.2 & \textbf{65.5} & 63.4 & 64.0 & \textbf{63.9} & 73.0 & 65.1 & \textbf{72.5} \\
\midrule
\multirow{4}{*}{Math}
& AIME24         & \textbf{4.7} & \textbf{4.7} & \textbf{4.7} & 29.3 & \textbf{34.7} & 33.3 & 38.0 & \textbf{46.7} & 45.3 \\
& AMC23          & 20.5 & 23.0 & \textbf{24.5} & 70.5 & 77.5 & \textbf{80.5} & 75.0 & \textbf{81.5} & 79.5 \\
& GSM8K          & 90.7 & 89.6 & \textbf{92.3} & 72.5 & \textbf{83.1} & 83.0 & 95.1 & \textbf{96.1} & 96.0 \\
& Math500        & 40.8 & 38.0 & \textbf{45.4} & 62.8 & 67.0 & \textbf{67.2} & 73.8 & \textbf{75.8} & \textbf{75.8} \\
\midrule
\multirow{2}{*}{Scientific QA}
& MMLU-Pro       & 42.3 & 43.2 & \textbf{47.8} & \textbf{53.6} & 53.4 & 54.1 & 63.7 & 68.7 & \textbf{69.1} \\
& GPQA           & 30.8 & 28.8 & \textbf{32.3} & 39.4 & \textbf{43.9} & 42.4 & 43.9 & \textbf{57.6} & 56.6 \\
\midrule
\multirow{3}{*}{Reasoning}
& MuSR           & 47.6 & \textbf{52.9} & 52.1 & \textbf{53.0} & \textbf{53.0} & 52.9 & 63.0 & 63.2 & \textbf{64.3} \\
& DROP           & 62.3 & 61.8 & \textbf{63.3} & \textbf{54.3} & 42.5 & 50.0 & 74.6 & \textbf{74.8} & 74.4 \\
& QASC           & 83.3 & 83.5 & \textbf{85.1} & 83.8 & \textbf{84.6} & 83.5 & 93.4 & 94.1 & \textbf{94.2} \\
\midrule
\multirow{2}{*}{QA}
& 2wiki          & 29.5 & 26.3 & \textbf{30.6} & 26.7 & 24.4 & \textbf{27.6} & 40.6 & \textbf{42.2} & 41.4 \\
& HotpotQA       & 28.6 & 28.1 & \textbf{29.1} & 18.5 & 17.3 & \textbf{19.5} & 27.7 & \textbf{29.5} & 28.6 \\
\multirow{2}{*}{QA-RAG}
& 2wiki\_RAG     & 34.1 & 33.2 & \textbf{34.3} & \textbf{36.5} & 24.7 & 29.2 & \textbf{56.0} & 55.6 & 55.1 \\
& HotpotQA\_RAG  & 31.0 & 31.4 & \textbf{31.9} & \textbf{27.1} & 21.7 & 23.9 & 39.3 & \textbf{39.6} & 38.9 \\
\bottomrule
\end{tabular}
\end{table*}

\subsection{Ablation Study}
\label{app:ablation}

We conduct thorough ablation experiments on each supervision dimension, for each model and each training dataset, as shown in \autoref{app:exp:abl-rb2-llama}, \autoref{app:exp:abl-rb2-dsllama}, \autoref{app:exp:abl-rb2-qwen}, \autoref{app:exp:abl-rag-llama}, \autoref{app:exp:abl-rag-dsllama} and \autoref{app:exp:abl-rag-qwen}. Across all settings, the results show a consistent trend: no single dimension is sufficient to yield robust improvements across diverse tasks. Combining multiple complementary dimensions produces cooperative effects that enhance generalization and no single dimension is dominant.

\begin{table}[t]
\centering
\caption{Ablation results of single dimension supervised training \textsc{LLaMA3.1-8B-Instruct} on RewardBench~2. Training pairs are selected from \textbf{\textsc{any}} subset. \textbf{DRM} means training with DRM supervision. All training pairs are selected from \textbf{\textsc{any}} subset.}
\label{app:exp:abl-rb2-llama}

\renewcommand{\arraystretch}{1.1}
\setlength{\tabcolsep}{3pt}
\scriptsize
\begin{tabular}{ll c>{\columncolor{mylighterblue}}c>{\columncolor{mylighterblue}}c>{\columncolor{mylighterblue}}c>{\columncolor{mylightblue}}c}
\toprule
\textbf{Task Domain} & \textbf{Dataset} & Native & Confidence & Coherence & Relevance & \textbf{DRM} \\
\midrule
\multirow{4}{*}{Code} 
& CodeMMLU       & 58.8 & 57.5 & 58.4 & 55.1 & \textbf{59.9} \\
& CodeScope      & 34.8 & 39.7 & 39.5 & 31.8 & \textbf{41.1} \\
& Cruxeval       & 50.4 & 53.9 & 53.5 & 32.4 & \textbf{57.5} \\
& Execution-v2   & 38.2 & 40.3 & 44.3 & 38.4 & \textbf{45.3} \\
\midrule
\multirow{2}{*}{Preference}
& RM-Bench       & 56.4 & 59.2 & 60.8 & 59.1 & \textbf{61.0} \\
& UltraFeedback  & 66.6 & 65.3 & 67.8 & 64.7 & \textbf{69.9} \\
\midrule
\multirow{4}{*}{Math}
& AIME24         & 4.7 & 4.0 & 4.7 & 2.7 &\textbf{ 6.0} \\
& AMC23          & 22.5 & 23.0 & 24.5 & 27.5 & \textbf{29.5} \\
& GSM8K          & 88.8 & 83.0 & 89.8 & 89.7 & \textbf{91.8} \\
& Math500        & 39.6 & 40.4 & 45.8 & 43.4 & \textbf{48.4} \\
\midrule
\multirow{2}{*}{Scientific QA}
& MMLU-Pro       & 41.9 & 46.7 & 48.7 & 42.0 & \textbf{48.7} \\
& GPQA           & 31.3 & 28.3 & 29.8 & 25.3 & \textbf{35.9} \\
\midrule
\multirow{3}{*}{Reasoning}
& MuSR           & 48.3 & 48.7 & 50.8 & 46.2 & \textbf{51.7} \\
& DROP           & 56.9 & 50.4 & \textbf{64.5} & 27.6 & 63.6 \\
& QASC           & 84.4 & 84.0 & 86.3 & 77.2 & \textbf{87.2} \\
\midrule
\multirow{2}{*}{QA}
& 2wiki          & 33.8 & 34.9 & 32.2 & 29.0 & \textbf{35.6} \\
& HotpotQA       & 29.3 & 29.6 & 30.0 & 26.0 & \textbf{31.8} \\
\multirow{2}{*}{QA-RAG}
& 2wiki\_RAG     & 31.2 & 28.5 & 36.1 & 31.2 & \textbf{39.9} \\
& HotpotQA\_RAG  & 28.3 & 27.1 & 33.1 & 27.4 & \textbf{34.5} \\
\bottomrule
\end{tabular}
\end{table}

\begin{table}[t]
\centering
\caption{Ablation results of single dimension supervised training \textsc{R1-distil-LLaMA8B} on RewardBench~2. Training pairs are selected from \textbf{\textsc{any}} subset. \textbf{DRM} means training with DRM supervision. All training pairs are selected from \textbf{\textsc{any}} subset.}
\label{app:exp:abl-rb2-dsllama}

\renewcommand{\arraystretch}{1.1}
\setlength{\tabcolsep}{3pt}
\scriptsize
\begin{tabular}{ll c>{\columncolor{mylighterblue}}c>{\columncolor{mylighterblue}}c>{\columncolor{mylighterblue}}c>{\columncolor{mylightblue}}c}
\toprule
\textbf{Task Domain} & \textbf{Dataset} & Native & Confidence & Coherence & Relevance & \textbf{DRM} \\
\midrule
\multirow{4}{*}{Code} 
& CodeMMLU       & 59.7 & 60.2 & 63.9 & 62.8 & \textbf{66.3} \\
& CodeScope      & 67.4 & 65.9 & 67.4 & 66.3 & \textbf{70.2} \\
& Cruxeval       & 71.9 & 73.5 & 76.1 & 73.4 & \textbf{77.2} \\
& Execution-v2   & 80.8 & 81.0 & 84.6 & 81.8 & \textbf{86.0} \\
\midrule
\multirow{2}{*}{Preference}
& RM-Bench       & 71.9 & 71.3 & 70.5 & 68.7 & \textbf{74.6} \\
& UltraFeedback  & 65.2 & 64.6 & 64.8 & 65.0 & \textbf{66.8} \\
\midrule
\multirow{4}{*}{Math}
& AIME24         & 28.7 & 26.0 & 29.3 & 31.3 & \textbf{33.3} \\
& AMC23          & 70.5 & 70.5 & 71.0 & 69.5 & \textbf{75.5} \\
& GSM8K          & 66.7 & 69.7 & 67.8 & \textbf{73.2} & 69.2 \\
& Math500        & 62.6 & 60.4 & 62.6 & \textbf{64.2} & 63.2 \\
\midrule
\multirow{2}{*}{Scientific QA}
& MMLU-Pro       & 51.5 & 52.3 & 52.9 & 51.6 & \textbf{54.7} \\
& GPQA           & 39.9 & 39.9 & 39.9 & 40.9 & \textbf{44.9} \\
\midrule
\multirow{3}{*}{Reasoning}
& MuSR           & 52.6 & 51.9 & \textbf{54.2} & 52.5 & 54.1 \\
& DROP           & 50.8 & 56.4 & \textbf{55.3} & 29.5 & 50.2 \\
& QASC           & 82.1 & 81.6 & 83.2 & 80.8 & \textbf{84.1} \\
\midrule
\multirow{2}{*}{QA}
& 2wiki          & 26.2 & 26.6 & 30.5 & 15.0 & \textbf{31.6} \\
& HotpotQA       & 18.1 & 17.8 & 19.1 & 13.6 & \textbf{19.7} \\
\multirow{2}{*}{QA-RAG}
& 2wiki\_RAG     & 36.7 & \textbf{39.3} & 39.1 & 20.2 & 37.9 \\
& HotpotQA\_RAG  & 27.1 & 27.1 &\textbf{ 27.9} & 17.9 & 27.3 \\
\bottomrule
\end{tabular}
\end{table}

\begin{table}[t]
\centering
\caption{
Ablation results of single dimension supervised training \textsc{Qwen3-8B} on RewardBench~2. Training pairs are selected from \textbf{\textsc{any}} subset. \textbf{DRM} means training with DRM supervision. All training pairs are selected from \textbf{\textsc{any}} subset.}
\label{app:exp:abl-rb2-qwen}

\renewcommand{\arraystretch}{1.1}
\setlength{\tabcolsep}{3pt}
\scriptsize
\begin{tabular}{ll c>{\columncolor{mylighterblue}}c>{\columncolor{mylighterblue}}c>{\columncolor{mylighterblue}}c>{\columncolor{mylightblue}}c}
\toprule
\textbf{Task Domain} & \textbf{Dataset} & Native & Confidence & Coherence & Relevance & \textbf{DRM} \\
\midrule
\multirow{4}{*}{Code} 
& CodeMMLU       & 77.9 & 78.0 & 79.5 & 78.1 & \textbf{80.3} \\
& CodeScope      & 86.5 & 86.8 & 86.3 & 86.7 & \textbf{87.4} \\
& Cruxeval       & 91.6 & 92.9 & 91.8 & 91.9 & \textbf{93.0} \\
& Execution-v2   & 98.5 & 98.1 & 98.3 & 98.3 & \textbf{99.0} \\
\midrule
\multirow{2}{*}{Preference}
& RM-Bench       & 85.4 & 84.8 & 84.6 & 84.8 & \textbf{85.6} \\
& UltraFeedback  & 71.3 & 71.1 & 72.0 & 72.0 & \textbf{73.2} \\
\midrule
\multirow{4}{*}{Math}
& AIME24         & 38.0 & 38.7 & 39.3 & \textbf{45.3} & 44.7 \\
& AMC23          & 72.0 & 72.5 & 77.5 & \textbf{79.5} & 79.0 \\
& GSM8K          & 95.6 & 95.2 & \textbf{95.7} & 95.4 & 96.1 \\
& Math500        & 73.2 & 72.4 & 74.6 & 74.2 & \textbf{75.6} \\
\midrule
\multirow{2}{*}{Scientific QA}
& MMLU-Pro       & 65.3 & 62.5 & 70.5 & 70.7 & \textbf{71.4} \\
& GPQA           & 48.0 & 43.9 & 54.0 & 56.1 & \textbf{58.1} \\
\midrule
\multirow{3}{*}{Reasoning}
& MuSR           & 63.5 & 64.8 & 63.5 & 62.8 & \textbf{65.5} \\
& DROP           & 74.7 & 74.4 & 74.0 & 74.6 & \textbf{74.9} \\
& QASC           & 94.1 & 93.8 & 93.7 & 94.0 & \textbf{94.2} \\
\midrule
\multirow{2}{*}{QA}
& 2wiki          & 39.8 & 40.9 & 39.5 & 42.0 & \textbf{42.2} \\
& HotpotQA       & 29.2 & 28.3 & 28.7 & 27.3 & \textbf{29.4} \\
\multirow{2}{*}{QA-RAG}
& 2wiki\_RAG     & 55.7 & 55.7 & 55.1 & 55.9 & \textbf{56.1} \\
& HotpotQA\_RAG  & 40.5 & 40.2 & 40.0 & 40.2 & \textbf{40.7} \\
\bottomrule
\end{tabular}
\end{table}

\begin{table}[t]
\centering
\caption{
Ablation results of single dimension supervised training \textsc{LLaMA3.1-8B-Instruct} on HotpotQA with RAG. Training pairs are selected from \textbf{\textsc{any}} subset. \textbf{DRM} means training with DRM supervision. All training pairs are selected from \textbf{\textsc{any}} subset.}
\label{app:exp:abl-rag-llama}

\renewcommand{\arraystretch}{1.1}
\setlength{\tabcolsep}{3pt}
\scriptsize
\begin{tabular}{ll c>{\columncolor{mylighterblue}}c>{\columncolor{mylighterblue}}c>{\columncolor{mylighterblue}}c>{\columncolor{mylightblue}}c}
\toprule
\textbf{Task Domain} & \textbf{Dataset} & Native & Confidence & Coherence & Relevance & \textbf{DRM} \\
\midrule
\multirow{4}{*}{Code} 
& CodeMMLU       & 58.8 & 58.7 & 59.6 & 58.1 & \textbf{60.5} \\
& CodeScope      & 34.8 & 37.6 & \textbf{41.8} & 39.3 & 41.7 \\
& Cruxeval       & 50.4 & 54.5 & 54.0 & 52.5 & \textbf{56.2} \\
& Execution-v2   & 38.2 & 42.8 & \textbf{43.6} & 39.7 & 43.4 \\
\midrule
\multirow{2}{*}{Preference}
& RM-Bench       & 56.4 & 59.9 & 60.3 & 59.4 & \textbf{62.9} \\
& UltraFeedback  & 66.6 & 65.6 & 66.4 & 65.3 & \textbf{68.2} \\
\midrule
\multirow{4}{*}{Math}
& AIME24         & 4.7 & 3.3 & \textbf{6.7} & 5.3 & 3.3 \\
& AMC23          & 22.5 & 22.5 & 26.0 & 19.5 & \textbf{28.5} \\
& GSM8K          & 88.8 & 87.8 & 89.6 & 90.6 & \textbf{91.5} \\
& Math500        & 39.6 & 41.2 & \textbf{46.2} & 41.8 & 45.0 \\
\midrule
\multirow{2}{*}{Scientific QA}
& MMLU-Pro       & 41.9 & 46.2 & 47.9 & 46.5 & \textbf{49.6} \\
& GPQA           & 31.3 & 28.3 & 29.3 & 31.8 & \textbf{34.3} \\
\midrule
\multirow{3}{*}{Reasoning}
& MuSR           & 48.3 & 48.0 & 51.6 & 50.7 & \textbf{53.0} \\
& DROP           & 56.9 & 62.3 & 65.3 & 58.9 & \textbf{67.3} \\
& QASC           & 84.4 & 83.4 & 85.4 & \textbf{87.5} & \textbf{87.5} \\
\midrule
\multirow{2}{*}{QA}
& 2wiki          & 33.8 & 35.2 & 38.4 & 37.6 & \textbf{40.9} \\
& HotpotQA       & 29.3 & 31.7 & \textbf{33.2} & 30.7 & 32.8 \\
\multirow{2}{*}{QA-RAG}
& 2wiki\_RAG     & 31.2 & 32.8 & 45.5 & 43.0 & \textbf{48.4} \\
& HotpotQA\_RAG  & 28.3 & 30.0 & 38.7 & 34.0 & \textbf{40.8} \\
\bottomrule
\end{tabular}
\end{table}

\begin{table}[t]
\centering
\caption{
Ablation results of single dimension supervised training \textsc{R1-distil-LLaMA8B} on HotpotQA with RAG. Training pairs are selected from \textbf{\textsc{any}} subset. \textbf{DRM} means training with DRM supervision. All training pairs are selected from \textbf{\textsc{any}} subset.}
\label{app:exp:abl-rag-dsllama}

\renewcommand{\arraystretch}{1.1}
\setlength{\tabcolsep}{3pt}
\scriptsize
\begin{tabular}{ll c>{\columncolor{mylighterblue}}c>{\columncolor{mylighterblue}}c>{\columncolor{mylighterblue}}c>{\columncolor{mylightblue}}c}
\toprule
\textbf{Task Domain} & \textbf{Dataset} & Native & Confidence & Coherence & Relevance & \textbf{DRM} \\
\midrule
\multirow{4}{*}{Code} 
& CodeMMLU       & 59.7 & 64.2 & 64.6 & 65.4 & \textbf{66.6} \\
& CodeScope      & 67.4 & 67.3 & 68.5 & 69.2 & \textbf{69.7} \\
& Cruxeval       & 71.9 & 71.9 & 75.2 & 73.1 & \textbf{75.4} \\
& Execution-v2   & 80.8 & 80.8 & 83.5 & 81.8 & \textbf{85.6} \\
\midrule
\multirow{2}{*}{Preference}
& RM-Bench       & 71.9 & 71.9 & 70.0 & 70.2 & \textbf{72.9} \\
& UltraFeedback  & 65.2 & 65.0 & 64.0 & 64.4 & \textbf{67.0} \\
\midrule
\multirow{4}{*}{Math}
& AIME24         & 28.7 & 28.0 & \textbf{32.7} & 30.7 & 30.7 \\
& AMC23          & 70.5 & 64.0 & 79.5 & 77.5 & \textbf{80.5} \\
& GSM8K          & 66.7 & 66.2 & 84.7 & \textbf{87.8} & 86.4 \\
& Math500        & 62.6 & 58.6 & 65.8 & 65.4 & \textbf{67.2} \\
\midrule
\multirow{2}{*}{Scientific QA}
& MMLU-Pro       & 51.5 & 52.4 & 51.6 & 53.5 & \textbf{54.9} \\
& GPQA           & 39.9 & \textbf{41.4 }& 39.9 & 38.4 & \textbf{41.4} \\
\midrule
\multirow{3}{*}{Reasoning}
& MuSR           & 52.6 & 54.8 & 56.0 & 57.5 & \textbf{58.6} \\
& DROP           & 50.8 & 63.0 & 63.0 & 41.0 & \textbf{65.4} \\
& QASC           & 82.1 & 84.6 & 84.1 & 83.1 & \textbf{85.2} \\
\midrule
\multirow{2}{*}{QA}
& 2wiki          & 26.2 & 24.3 & 37.3 & 7.1 & \textbf{37.9} \\
& HotpotQA       & 18.1 & 20.2 & 22.6 & 13.9 & \textbf{24.0} \\
\multirow{2}{*}{QA-RAG}
& 2wiki\_RAG     & 36.7 & 46.0 & 49.6 & 25.6 & \textbf{51.6} \\
& HotpotQA\_RAG  & 27.1 & 33.5 & 35.8 & 22.6 & \textbf{37.2} \\
\bottomrule
\end{tabular}
\end{table}

\begin{table}[t]
\centering
\caption{Ablation results of single dimension supervised training \textsc{Qwen3-8B}on HotpotQA with RAG. Training pairs are selected from \textbf{\textsc{any}} subset. \textbf{DRM} means training with DRM supervision. All training pairs are selected from \textbf{\textsc{any}} subset.}
\label{app:exp:abl-rag-qwen}

\renewcommand{\arraystretch}{1.1}
\setlength{\tabcolsep}{3pt}
\scriptsize
\begin{tabular}{ll c>{\columncolor{mylighterblue}}c>{\columncolor{mylighterblue}}c>{\columncolor{mylighterblue}}c>{\columncolor{mylightblue}}c}
\toprule
\textbf{Task Domain} & \textbf{Dataset} & Native & Confidence & Coherence & Relevance & \textbf{DRM} \\
\midrule
\multirow{4}{*}{Code} 
& CodeMMLU       & 77.9 & 77.3 & 79.0 & \textbf{79.2} & 79.0 \\
& CodeScope      & 86.5 & 86.6 & 86.6 & 86.8 & \textbf{87.3} \\
& Cruxeval       & 91.6 & 92.1 & 92.0 & 91.9 & \textbf{92.2} \\
& Execution-v2   & 98.5 & 98.1 & 97.9 & 98.3 & \textbf{98.7} \\
\midrule
\multirow{2}{*}{Preference}
& RM-Bench       & 85.4 & \textbf{85.9} & 84.6 & 83.8 & 85.2 \\
& UltraFeedback  & 71.3 & 72.0 & 71.4 & 72.2 & \textbf{72.7} \\
\midrule
\multirow{4}{*}{Math}
& AIME24         & 38.0 & 38.0 & 44.0 & 45.3 & \textbf{47.3} \\
& AMC23          & 72.0 & 73.0 & 78.5 & \textbf{83.5} & 82.5 \\
& GSM8K          & 95.6 & 95.5 & 95.4 & 95.7 & \textbf{96.0} \\
& Math500        & 73.2 & 72.0 & 75.6 & 76.2 & \textbf{76.4} \\
\midrule
\multirow{2}{*}{Scientific QA}
& MMLU-Pro       & 65.3 & 62.4 & \textbf{71.1} & 70.1 & 70.4 \\
& GPQA           & 48.0 & 42.9 & 52.0 & 55.6 & \textbf{56.1} \\
\midrule
\multirow{3}{*}{Reasoning}
& MuSR           & 63.5 & \textbf{63.9} & 62.7 & 63.2 & 63.5 \\
& DROP           & \textbf{74.7} & 73.2 & 73.5 & 74.4 & 74.2 \\
& QASC           & \textbf{94.1} & 92.7 & 93.7 & 94.0 & 94.0 \\
\midrule
\multirow{2}{*}{QA}
& 2wiki          & 39.8 & \textbf{40.9} & 38.9 & 37.5 & 40.1 \\
& HotpotQA       & \textbf{29.2} & 27.2 & 28.5 & 28.2 & 28.7 \\
\multirow{2}{*}{QA-RAG}
& 2wiki\_RAG     & 55.7 & 55.5 & 55.5 & 53.9 & \textbf{56.9} \\
& HotpotQA\_RAG  & \textbf{40.5} & 38.7 & 38.1 & 35.4 & 38.8 \\
\bottomrule
\end{tabular}
\end{table}

\end{document}

%% file: math_commands.tex
\usepackage{amsmath,amsfonts,bm}

\def\eqref#1{equation~\ref{#1}}
\def\1{\bm{1}}

\DeclareMathAlphabet{\mathsfit}{\encodingdefault}{\sfdefault}{m}{sl}
\SetMathAlphabet{\mathsfit}{bold}{\encodingdefault}{\sfdefault}{bx}{n}

%% file: iclr2026_conference.bbl
\begin{thebibliography}{59}
\providecommand{\natexlab}[1]{#1}
\providecommand{\url}[1]{\texttt{#1}}
\expandafter\ifx\csname urlstyle\endcsname\relax
  \providecommand{\doi}[1]{doi: #1}\else
  \providecommand{\doi}{doi: \begingroup \urlstyle{rm}\Url}\fi

\bibitem[Chen et~al.(2025)Chen, Li, Wang, Jin, Qian, Wang, Wang, Zhang, Zhang, Zhang, Tong, and Ji]{RMR1Reward2025}
Xiusi Chen, Gaotang Li, Ziqi Wang, Bowen Jin, Cheng Qian, Yu~Wang, Hongru Wang, Yu~Zhang, Denghui Zhang, Tong Zhang, Hanghang Tong, and Heng Ji.
\newblock {{RM-R1}}: {{Reward Modeling}} as {{Reasoning}}, May 2025.

\bibitem[Cheng et~al.(2025)Cheng, Qiao, Li, Guo, Wang, Xiong, Lv, and Wang]{StopSummation2025}
Jie Cheng, Ruixi Qiao, Lijun Li, Chao Guo, Junle Wang, Gang Xiong, Yisheng Lv, and Fei-Yue Wang.
\newblock Stop {{Summation}}: {{Min-Form Credit Assignment Is All Process Reward Model Needs}} for {{Reasoning}}, May 2025.

\bibitem[Christiano et~al.(2023)Christiano, Leike, Brown, Martic, Legg, and Amodei]{DeepReinforcement2023}
Paul Christiano, Jan Leike, Tom~B. Brown, Miljan Martic, Shane Legg, and Dario Amodei.
\newblock Deep reinforcement learning from human preferences, February 2023.

\bibitem[Cobbe et~al.(2021{\natexlab{a}})Cobbe, Kosaraju, Bavarian, Chen, Jun, Kaiser, Plappert, Tworek, Hilton, Nakano, Hesse, and Schulman]{TrainingVerifiers2021}
Karl Cobbe, Vineet Kosaraju, Mohammad Bavarian, Mark Chen, Heewoo Jun, Lukasz Kaiser, Matthias Plappert, Jerry Tworek, Jacob Hilton, Reiichiro Nakano, Christopher Hesse, and John Schulman.
\newblock Training {{Verifiers}} to {{Solve Math Word Problems}}, November 2021{\natexlab{a}}.

\bibitem[Cobbe et~al.(2021{\natexlab{b}})Cobbe, Kosaraju, Bavarian, Chen, Jun, Kaiser, Plappert, Tworek, Hilton, Nakano, Hesse, and Schulman]{cobbe2021GSM8K}
Karl Cobbe, Vineet Kosaraju, Mohammad Bavarian, Mark Chen, Heewoo Jun, Lukasz Kaiser, Matthias Plappert, Jerry Tworek, Jacob Hilton, Reiichiro Nakano, Christopher Hesse, and John Schulman.
\newblock Training verifiers to solve math word problems.
\newblock \emph{arXiv preprint arXiv:2110.14168}, 2021{\natexlab{b}}.

\bibitem[Cui et~al.(2024)Cui, Yuan, Ding, Yao, He, Zhu, Ni, Xie, Xie, Lin, Liu, and Sun]{UltraFeedbackBoosting2024}
Ganqu Cui, Lifan Yuan, Ning Ding, Guanming Yao, Bingxiang He, Wei Zhu, Yuan Ni, Guotong Xie, Ruobing Xie, Yankai Lin, Zhiyuan Liu, and Maosong Sun.
\newblock {{UltraFeedback}}: {{Boosting Language Models}} with {{Scaled AI Feedback}}, July 2024.

\bibitem[Cui et~al.(2025)Cui, Zhang, Chen, Yuan, Wang, Zuo, Li, Fan, Chen, Chen, Liu, Peng, Bai, Ouyang, Cheng, Zhou, and Ding]{EntropyMechanism2025}
Ganqu Cui, Yuchen Zhang, Jiacheng Chen, Lifan Yuan, Zhi Wang, Yuxin Zuo, Haozhan Li, Yuchen Fan, Huayu Chen, Weize Chen, Zhiyuan Liu, Hao Peng, Lei Bai, Wanli Ouyang, Yu~Cheng, Bowen Zhou, and Ning Ding.
\newblock The {{Entropy Mechanism}} of {{Reinforcement Learning}} for {{Reasoning Language Models}}, May 2025.

\bibitem[{DeepSeek-AI} et~al.(2025){DeepSeek-AI}, Guo, Yang, Zhang, Song, Zhang, Xu, Zhu, Ma, Wang, Bi, Zhang, Yu, Wu, Wu, Gou, Shao, Li, Gao, Liu, Xue, Wang, Wu, Feng, and et~al]{DeepSeekR1Incentivizing2025}
{DeepSeek-AI}, Daya Guo, Dejian Yang, Haowei Zhang, Junxiao Song, Ruoyu Zhang, Runxin Xu, Qihao Zhu, Shirong Ma, Peiyi Wang, Xiao Bi, Xiaokang Zhang, Xingkai Yu, Yu~Wu, Z.~F. Wu, Zhibin Gou, Zhihong Shao, Zhuoshu Li, Ziyi Gao, Aixin Liu, Bing Xue, Bingxuan Wang, Bochao Wu, Bei Feng, and et~al.
\newblock {{DeepSeek-R1}}: {{Incentivizing Reasoning Capability}} in {{LLMs}} via {{Reinforcement Learning}}, January 2025.

\bibitem[Dua et~al.(2019)Dua, Wang, Dasigi, Stanovsky, Singh, and Gardner]{Dua2019DROP}
Dheeru Dua, Yizhong Wang, Pradeep Dasigi, Gabriel Stanovsky, Sameer Singh, and Matt Gardner.
\newblock {DROP}: A reading comprehension benchmark requiring discrete reasoning over paragraphs.
\newblock In \emph{Proc. of NAACL}, 2019.

\bibitem[Golovneva et~al.(2023)Golovneva, Chen, Poff, Corredor, Zettlemoyer, {Fazel-Zarandi}, and Celikyilmaz]{ROSCOESuite2023}
Olga Golovneva, Moya Chen, Spencer Poff, Martin Corredor, Luke Zettlemoyer, Maryam {Fazel-Zarandi}, and Asli Celikyilmaz.
\newblock {{ROSCOE}}: {{A Suite}} of {{Metrics}} for {{Scoring Step-by-Step Reasoning}}, September 2023.

\bibitem[Grattafiori et~al.(2024)Grattafiori, Dubey, Jauhri, Pandey, Kadian, {Al-Dahle}, Letman, Mathur, Schelten, Vaughan, Yang, Fan, Goyal, Hartshorn, Yang, Mitra, Sravankumar, Korenev, Hinsvark, Rao, Zhang, Rodriguez, Gregerson, Spataru, Roziere, Biron, Tang, Chern, Caucheteux, and et~al]{Llama32024}
Aaron Grattafiori, Abhimanyu Dubey, Abhinav Jauhri, Abhinav Pandey, Abhishek Kadian, Ahmad {Al-Dahle}, Aiesha Letman, Akhil Mathur, Alan Schelten, Alex Vaughan, Amy Yang, Angela Fan, Anirudh Goyal, Anthony Hartshorn, Aobo Yang, Archi Mitra, Archie Sravankumar, Artem Korenev, Arthur Hinsvark, Arun Rao, Aston Zhang, Aurelien Rodriguez, Austen Gregerson, Ava Spataru, Baptiste Roziere, Bethany Biron, Binh Tang, Bobbie Chern, Charlotte Caucheteux, and et~al.
\newblock The {{Llama}} 3 {{Herd}} of {{Models}}, November 2024.

\bibitem[Gu et~al.(2024)Gu, Rozière, Leather, Solar-Lezama, Synnaeve, and Wang]{gu2024Cruxeval}
Alex Gu, Baptiste Rozière, Hugh Leather, Armando Solar-Lezama, Gabriel Synnaeve, and Sida~I. Wang.
\newblock Cruxeval: A benchmark for code reasoning, understanding and execution.
\newblock \emph{arXiv preprint arXiv:2401.03065}, 2024.

\bibitem[Ho et~al.(2020)Ho, Nguyen, Sugawara, and Aizawa]{ConstructingMultihop2020}
Xanh Ho, Anh-Khoa~Duong Nguyen, Saku Sugawara, and Akiko Aizawa.
\newblock Constructing {{A Multi-hop QA Dataset}} for {{Comprehensive Evaluation}} of {{Reasoning Steps}}, November 2020.

\bibitem[Jain et~al.(2024)Jain, Han, Gu, Li, Yan, Zhang, Wang, {Solar-Lezama}, Sen, and Stoica]{LiveCodeBenchHolistic2024}
Naman Jain, King Han, Alex Gu, Wen-Ding Li, Fanjia Yan, Tianjun Zhang, Sida Wang, Armando {Solar-Lezama}, Koushik Sen, and Ion Stoica.
\newblock {{LiveCodeBench}}: {{Holistic}} and {{Contamination Free Evaluation}} of {{Large Language Models}} for {{Code}}, June 2024.

\bibitem[Jin et~al.(2024)Jin, Zhu, Yang, Zhang, and Dou]{FlashRAG}
Jiajie Jin, Yutao Zhu, Xinyu Yang, Chenghao Zhang, and Zhicheng Dou.
\newblock Flashrag: A modular toolkit for efficient retrieval-augmented generation research.
\newblock \emph{CoRR}, abs/2405.13576, 2024.
\newblock URL \url{https://arxiv.org/abs/2405.13576}.

\bibitem[Khot et~al.(2020)Khot, Clark, Guerquin, Jansen, and Sabharwal]{allenai:QASC}
Tushar Khot, Peter Clark, Michal Guerquin, Peter Jansen, and Ashish Sabharwal.
\newblock Qasc: A dataset for question answering via sentence composition.
\newblock \emph{arXiv:1910.11473v2}, 2020.

\bibitem[Kwon et~al.(2023)Kwon, Li, Zhuang, Sheng, Zheng, Yu, Gonzalez, Zhang, and Stoica]{kwon2023efficient}
Woosuk Kwon, Zhuohan Li, Siyuan Zhuang, Ying Sheng, Lianmin Zheng, Cody~Hao Yu, Joseph~E. Gonzalez, Hao Zhang, and Ion Stoica.
\newblock Efficient memory management for large language model serving with pagedattention.
\newblock In \emph{Proceedings of the ACM SIGOPS 29th Symposium on Operating Systems Principles}, 2023.

\bibitem[Kydlíček(2024)]{math-verify}
Hynek Kydlíček.
\newblock Math-verify: Math verification library.
\newblock \url{https://github.com/huggingface/math-verify}, 2024.
\newblock Version 0.6.1, Apache-2.0 License. Math-Verify is a library to rule-based verify mathematical answers.

\bibitem[Lambert et~al.(2025)Lambert, Morrison, Pyatkin, Huang, Ivison, Brahman, Miranda, Liu, Dziri, Lyu, Gu, Malik, Graf, Hwang, Yang, Bras, Tafjord, Wilhelm, Soldaini, Smith, Wang, Dasigi, and Hajishirzi]{Tulu32025}
Nathan Lambert, Jacob Morrison, Valentina Pyatkin, Shengyi Huang, Hamish Ivison, Faeze Brahman, Lester James~V. Miranda, Alisa Liu, Nouha Dziri, Shane Lyu, Yuling Gu, Saumya Malik, Victoria Graf, Jena~D. Hwang, Jiangjiang Yang, Ronan~Le Bras, Oyvind Tafjord, Chris Wilhelm, Luca Soldaini, Noah~A. Smith, Yizhong Wang, Pradeep Dasigi, and Hannaneh Hajishirzi.
\newblock Tulu 3: {{Pushing Frontiers}} in {{Open Language Model Post-Training}}, April 2025.

\bibitem[Leang et~al.(2025)Leang, Zhao, Gema, Yang, Kwan, He, Li, Minervini, Giunchiglia, and Cohen]{PiCSARProbabilistic2025}
Joshua Ong~Jun Leang, Zheng Zhao, Aryo~Pradipta Gema, Sohee Yang, Wai-Chung Kwan, Xuanli He, Wenda Li, Pasquale Minervini, Eleonora Giunchiglia, and Shay~B. Cohen.
\newblock {{PiCSAR}}: {{Probabilistic Confidence Selection And Ranking}} for {{Reasoning Chains}}, 2025.
\newblock URL \url{http://arxiv.org/abs/2508.21787}.

\bibitem[Lightman et~al.(2024)Lightman, Kosaraju, Burda, Edwards, Baker, Lee, Leike, Schulman, Sutskever, and Cobbe]{LETSVERIFY2024}
Hunter Lightman, Vineet Kosaraju, Yura Burda, Harri Edwards, Bowen Baker, Teddy Lee, Jan Leike, John Schulman, Ilya Sutskever, and Karl Cobbe.
\newblock {{LET}}'{{S VERIFY STEP BY STEP}}.
\newblock 2024.

\bibitem[Liu et~al.(2025)Liu, Zeng, Xiao, He, Liu, Wang, Yan, Shen, Zhang, Xu, Liu, and Zhou]{SkyworkRewardV2Scaling2025}
Chris~Yuhao Liu, Liang Zeng, Yuzhen Xiao, Jujie He, Jiacai Liu, Chaojie Wang, Rui Yan, Wei Shen, Fuxiang Zhang, Jiacheng Xu, Yang Liu, and Yahui Zhou.
\newblock Skywork-{{Reward-V2}}: {{Scaling Preference Data Curation}} via {{Human-AI Synergy}}, July 2025.

\bibitem[Liu et~al.(2024{\natexlab{a}})Liu, Lei, Wang, Huang, Feng, Wen, Cheng, Ke, Xu, Tam, Zhang, Sun, Gu, Wang, Zhang, Huang, Dong, and Tang]{AlignBenchBenchmarking2024}
Xiao Liu, Xuanyu Lei, Shengyuan Wang, Yue Huang, Zhuoer Feng, Bosi Wen, Jiale Cheng, Pei Ke, Yifan Xu, Weng~Lam Tam, Xiaohan Zhang, Lichao Sun, Xiaotao Gu, Hongning Wang, Jing Zhang, Minlie Huang, Yuxiao Dong, and Jie Tang.
\newblock {{AlignBench}}: {{Benchmarking Chinese Alignment}} of {{Large Language Models}}, August 2024{\natexlab{a}}.

\bibitem[Liu et~al.(2024{\natexlab{b}})Liu, Yao, Min, Cao, Hou, and Li]{RMBenchBenchmarking2024}
Yantao Liu, Zijun Yao, Rui Min, Yixin Cao, Lei Hou, and Juanzi Li.
\newblock {{RM-Bench}}: {{Benchmarking Reward Models}} of {{Language Models}} with {{Subtlety}} and {{Style}}, October 2024{\natexlab{b}}.

\bibitem[Luo et~al.(2024)Luo, Liu, Liu, Phatale, Guo, Lara, Li, Shu, Zhu, Meng, Sun, and Rastogi]{ImproveMathematical2024}
Liangchen Luo, Yinxiao Liu, Rosanne Liu, Samrat Phatale, Meiqi Guo, Harsh Lara, Yunxuan Li, Lei Shu, Yun Zhu, Lei Meng, Jiao Sun, and Abhinav Rastogi.
\newblock Improve {{Mathematical Reasoning}} in {{Language Models}} by {{Automated Process Supervision}}, December 2024.

\bibitem[Manh et~al.(2025)Manh, Chau, Hai, Doan, Nguyen, Pham, and Bui]{CodeMMLUMultiTask2025}
Dung~Nguyen Manh, Thang~Phan Chau, Nam~Le Hai, Thong~T. Doan, Nam~V. Nguyen, Quang Pham, and Nghi D.~Q. Bui.
\newblock {{CodeMMLU}}: {{A Multi-Task Benchmark}} for {{Assessing Code Understanding}} \& {{Reasoning Capabilities}} of {{CodeLLMs}}, April 2025.

\bibitem[OpenAI et~al.(2024)OpenAI, Jaech, Kalai, Lerer, Richardson, {El-Kishky}, Low, Helyar, Madry, Beutel, Carney, Iftimie, Karpenko, Passos, Neitz, Prokofiev, Wei, Tam, Bennett, Kumar, Saraiva, Vallone, Duberstein, Kondrich, Mishchenko, Applebaum, Jiang, Nair, Zoph, Ghorbani, Rossen, Sokolowsky, Barak, McGrew, Minaiev, Hao, Baker, Houghton, McKinzie, Eastman, and et~al]{OpenAIO12024}
OpenAI, Aaron Jaech, Adam Kalai, Adam Lerer, Adam Richardson, Ahmed {El-Kishky}, Aiden Low, Alec Helyar, Aleksander Madry, Alex Beutel, Alex Carney, Alex Iftimie, Alex Karpenko, Alex~Tachard Passos, Alexander Neitz, Alexander Prokofiev, Alexander Wei, Allison Tam, Ally Bennett, Ananya Kumar, Andre Saraiva, Andrea Vallone, Andrew Duberstein, Andrew Kondrich, Andrey Mishchenko, Andy Applebaum, Angela Jiang, Ashvin Nair, Barret Zoph, Behrooz Ghorbani, Ben Rossen, Benjamin Sokolowsky, Boaz Barak, Bob McGrew, Borys Minaiev, Botao Hao, Bowen Baker, Brandon Houghton, Brandon McKinzie, Brydon Eastman, and et~al.
\newblock {{OpenAI}} o1 {{System Card}}, December 2024.

\bibitem[Prasad et~al.(2023)Prasad, Saha, Zhou, and Bansal]{ReCEvalEvaluating2023}
Archiki Prasad, Swarnadeep Saha, Xiang Zhou, and Mohit Bansal.
\newblock {{ReCEval}}: {{Evaluating Reasoning Chains}} via {{Correctness}} and {{Informativeness}}, November 2023.

\bibitem[Rafailov et~al.(2023)Rafailov, Sharma, Mitchell, Manning, Ermon, and Finn]{DirectPreference}
Rafael Rafailov, Archit Sharma, Eric Mitchell, Christopher~D Manning, Stefano Ermon, and Chelsea Finn.
\newblock Direct preference optimization: {{Your}} language model is secretly a reward model.
\newblock In A.~Oh, T.~Naumann, A.~Globerson, K.~Saenko, M.~Hardt, and S.~Levine (eds.), \emph{Advances in Neural Information Processing Systems}, volume~36, pp.\  53728--53741. Curran Associates, Inc., 2023.

\bibitem[Rein et~al.(2023)Rein, Hou, Stickland, Petty, Pang, Dirani, Michael, and Bowman]{GPQAGraduateLevel2023}
David Rein, Betty~Li Hou, Asa~Cooper Stickland, Jackson Petty, Richard~Yuanzhe Pang, Julien Dirani, Julian Michael, and Samuel~R. Bowman.
\newblock {{GPQA}}: {{A Graduate-Level Google-Proof Q}}\&{{A Benchmark}}, November 2023.

\bibitem[Schulman et~al.(2017)Schulman, Wolski, Dhariwal, Radford, and Klimov]{ProximalPolicy2017}
John Schulman, Filip Wolski, Prafulla Dhariwal, Alec Radford, and Oleg Klimov.
\newblock Proximal {{Policy Optimization Algorithms}}, August 2017.

\bibitem[Shao et~al.(2024)Shao, Wang, Zhu, Xu, Song, Bi, Zhang, Zhang, Li, Wu, and Guo]{DeepSeekMathPushing2024}
Zhihong Shao, Peiyi Wang, Qihao Zhu, Runxin Xu, Junxiao Song, Xiao Bi, Haowei Zhang, Mingchuan Zhang, Y.~K. Li, Y.~Wu, and Daya Guo.
\newblock {{DeepSeekMath}}: {{Pushing}} the {{Limits}} of {{Mathematical Reasoning}} in {{Open Language Models}}, April 2024.

\bibitem[Sprague et~al.(2024)Sprague, Ye, Bostrom, Chaudhuri, and Durrett]{MuSRTesting2024}
Zayne Sprague, Xi~Ye, Kaj Bostrom, Swarat Chaudhuri, and Greg Durrett.
\newblock {{MuSR}}: {{Testing}} the {{Limits}} of {{Chain-of-thought}} with {{Multistep Soft Reasoning}}, March 2024.

\bibitem[Su et~al.(2025)Su, Pan, Bai, Liu, Dong, Huang, Hu, Zhang, Gai, and Zhou]{KlearReasonerAdvancing2025}
Zhenpeng Su, Leiyu Pan, Xue Bai, Dening Liu, Guanting Dong, Jiaming Huang, Wenping Hu, Fuzheng Zhang, Kun Gai, and Guorui Zhou.
\newblock Klear-{{Reasoner}}: {{Advancing Reasoning Capability}} via {{Gradient-Preserving Clipping Policy Optimization}}, August 2025.

\bibitem[Team et~al.(2025)Team, Zeng, Lv, Zheng, Hou, Chen, Xie, Wang, Yin, Zeng, Zhang, Wang, Zhong, Liu, Lu, Cao, Zhang, Huang, Wei, Cheng, and et~al]{GLM45Agentic2025}
GLM-4~5 Team, Aohan Zeng, Xin Lv, Qinkai Zheng, Zhenyu Hou, Bin Chen, Chengxing Xie, Cunxiang Wang, Da~Yin, Hao Zeng, Jiajie Zhang, Kedong Wang, Lucen Zhong, Mingdao Liu, Rui Lu, Shulin Cao, Xiaohan Zhang, Xuancheng Huang, Yao Wei, Yean Cheng, and et~al.
\newblock {{GLM-4}}.5: {{Agentic}}, {{Reasoning}}, and {{Coding}} ({{ARC}}) {{Foundation Models}}, August 2025.

\bibitem[von Werra et~al.(2020)von Werra, Belkada, Tunstall, Beeching, Thrush, Lambert, Huang, Rasul, and Gallouédec]{vonwerra2022trl}
Leandro von Werra, Younes Belkada, Lewis Tunstall, Edward Beeching, Tristan Thrush, Nathan Lambert, Shengyi Huang, Kashif Rasul, and Quentin Gallouédec.
\newblock Trl: Transformer reinforcement learning.
\newblock \url{https://github.com/huggingface/trl}, 2020.

\bibitem[Wang et~al.(2024)Wang, Ma, Zhang, Ni, Chandra, Guo, Ren, Arulraj, He, Jiang, Li, Ku, Wang, Zhuang, Fan, Yue, and Chen]{MMLUProMore2024}
Yubo Wang, Xueguang Ma, Ge~Zhang, Yuansheng Ni, Abhranil Chandra, Shiguang Guo, Weiming Ren, Aaran Arulraj, Xuan He, Ziyan Jiang, Tianle Li, Max Ku, Kai Wang, Alex Zhuang, Rongqi Fan, Xiang Yue, and Wenhu Chen.
\newblock {{MMLU-Pro}}: {{A More Robust}} and {{Challenging Multi-Task Language Understanding Benchmark}}, November 2024.

\bibitem[Wang et~al.()Wang, Zeng, Delalleau, Shin, Soares, Bukharin, Evans, Dong, and Kuchaiev]{HelpSteer3PreferenceOpen2025}
Zhilin Wang, Jiaqi Zeng, Olivier Delalleau, Hoo-Chang Shin, Felipe Soares, Alexander Bukharin, Ellie Evans, Yi~Dong, and Oleksii Kuchaiev.
\newblock {{HelpSteer3-Preference}}: {{Open Human-Annotated Preference Data}} across {{Diverse Tasks}} and {{Languages}}.
\newblock URL \url{http://arxiv.org/abs/2505.11475}.

\bibitem[Wang et~al.(2025)Wang, Dong, Luo, Ruan, Cheng, Chen, Li, and Liu]{wang2025multimodal}
Ziyue Wang, Yurui Dong, Fuwen Luo, Minyuan Ruan, Zhili Cheng, Chi Chen, Peng Li, and Yang Liu.
\newblock How do multimodal large language models handle complex multimodal reasoning? placing them in an extensible escape game.
\newblock \emph{arXiv e-prints}, pp.\  arXiv--2503, 2025.

\bibitem[Wu et~al.(2025)Wu, Chang, Liu, He, Xian, Hong, Chen, Tian, Yang, Shi, Lin, Yao, and Xu]{WeChatYATTScalable2025}
Junyu Wu, Weiming Chang, Xiaotao Liu, Guanyou He, Tingfeng Xian, Haoqiang Hong, Boqi Chen, Hongtao Tian, Tao Yang, Yunsheng Shi, Feng Lin, Ting Yao, and Jiatao Xu.
\newblock {{WeChat-YATT}}: {{A Scalable}}, {{Simple}}, {{Efficient}}, and {{Production Ready Training Library}}, August 2025.

\bibitem[Xie et~al.(2025)Xie, Shi, Tian, Yao, and Zhang]{CAPOEnhancing2025}
Guofu Xie, Yunsheng Shi, Hongtao Tian, Ting Yao, and Xiao Zhang.
\newblock {{CAPO}}: {{Towards Enhancing LLM Reasoning}} through {{Verifiable Generative Credit Assignment}}, August 2025.

\bibitem[Xiong et~al.(2025)Xiong, Zhao, Golovneva, Zhang, Weston, and Sukhbaatar]{StepWiserStepwise2025}
Wei Xiong, Wenting Zhao, Olga Golovneva, Tong Zhang, Jason Weston, and Sainbayar Sukhbaatar.
\newblock {{StepWiser}}: {{Stepwise Generative Judges}} for {{Wiser Reasoning}}, August 2025.

\bibitem[Xu et~al.(2025)Xu, Hao, Zong, Wang, Zhang, Wang, Lan, Gong, Ouyang, Meng, Shao, Yan, Yang, Song, Ren, Hu, Li, Feng, Gao, and Li]{LargeReasoning2025}
Fengli Xu, Qianyue Hao, Zefang Zong, Jingwei Wang, Yunke Zhang, Jingyi Wang, Xiaochong Lan, Jiahui Gong, Tianjian Ouyang, Fanjin Meng, Chenyang Shao, Yuwei Yan, Qinglong Yang, Yiwen Song, Sijian Ren, Xinyuan Hu, Yu~Li, Jie Feng, Chen Gao, and Yong Li.
\newblock Towards {{Large Reasoning Models}}: {{A Survey}} of {{Reinforced Reasoning}} with {{Large Language Models}}, January 2025.

\bibitem[Yan et~al.(2024)Yan, Liu, Wang, Li, Chen, Wang, Lin, Zhao, Zhu, Sundaram, and Deng]{CodeScopeExecutionbased2024}
Weixiang Yan, Haitian Liu, Yunkun Wang, Yunzhe Li, Qian Chen, Wen Wang, Tingyu Lin, Weishan Zhao, Li~Zhu, Hari Sundaram, and Shuiguang Deng.
\newblock {{CodeScope}}: {{An Execution-based Multilingual Multitask Multidimensional Benchmark}} for {{Evaluating LLMs}} on {{Code Understanding}} and {{Generation}}, June 2024.

\bibitem[Yang et~al.(2024)Yang, Yang, Hui, Zheng, Yu, Zhou, Li, Li, Liu, Huang, et~al.]{yang2024qwen2}
An~Yang, Baosong Yang, Binyuan Hui, Bo~Zheng, Bowen Yu, Chang Zhou, Chengpeng Li, Chengyuan Li, Dayiheng Liu, Fei Huang, et~al.
\newblock Qwen2 technical report.
\newblock \emph{arXiv preprint arXiv:2407.10671}, 2024.

\bibitem[Yang et~al.(2025)Yang, Li, Yang, Zhang, Hui, Zheng, Yu, Gao, Huang, Lv, Zheng, Liu, Zhou, Huang, Hu, Ge, Wei, Lin, Tang, Yang, Tu, Zhang, Yang, Yang, Zhou, Zhou, Lin, Dang, Bao, Yang, Yu, Deng, Li, Xue, Li, Zhang, Wang, Zhu, Men, Gao, Liu, Luo, Li, Tang, Yin, Ren, Wang, Zhang, Ren, Fan, Su, Zhang, Zhang, Wan, Liu, Wang, Cui, Zhang, Zhou, and Qiu]{Qwen3Technical2025}
An~Yang, Anfeng Li, Baosong Yang, Beichen Zhang, Binyuan Hui, Bo~Zheng, Bowen Yu, Chang Gao, Chengen Huang, Chenxu Lv, Chujie Zheng, Dayiheng Liu, Fan Zhou, Fei Huang, Feng Hu, Hao Ge, Haoran Wei, Huan Lin, Jialong Tang, Jian Yang, Jianhong Tu, Jianwei Zhang, Jianxin Yang, Jiaxi Yang, Jing Zhou, Jingren Zhou, Junyang Lin, Kai Dang, Keqin Bao, Kexin Yang, Le~Yu, Lianghao Deng, Mei Li, Mingfeng Xue, Mingze Li, Pei Zhang, Peng Wang, Qin Zhu, Rui Men, Ruize Gao, Shixuan Liu, Shuang Luo, Tianhao Li, Tianyi Tang, Wenbiao Yin, Xingzhang Ren, Xinyu Wang, Xinyu Zhang, Xuancheng Ren, Yang Fan, Yang Su, Yichang Zhang, Yinger Zhang, Yu~Wan, Yuqiong Liu, Zekun Wang, Zeyu Cui, Zhenru Zhang, Zhipeng Zhou, and Zihan Qiu.
\newblock Qwen3 {{Technical Report}}, May 2025.

\bibitem[Yang et~al.()Yang, Sun, Xin, Sun, Bhalla, Chen, Gui, Jiang, Jiang, Kong, Moran, Wang, Xu, Yan, Yang, Yuan, Zha, Tang, Chen, Scheffer, Liu, Shah, Wanga, Kumar, Yih, and Dong]{CRAGComprehensive}
Xiao Yang, Kai Sun, Hao Xin, Yushi Sun, Nikita Bhalla, Xiangsen Chen, Rongze~Daniel Gui, Ziran~Will Jiang, Ziyu Jiang, Lingkun Kong, Brian Moran, Jiaqi Wang, Yifan~Ethan Xu, An~Yan, Chenyu Yang, Eting Yuan, Hanwen Zha, Nan Tang, Lei Chen, Nicolas Scheffer, Yue Liu, Nirav Shah, Rakesh Wanga, Anuj Kumar, Wen-tau Yih, and Xin~Luna Dong.
\newblock {{CRAG}} -- {{Comprehensive RAG Benchmark}}.

\bibitem[Yang et~al.(2018)Yang, Qi, Zhang, Bengio, Cohen, Salakhutdinov, and Manning]{HotpotQADataset2018}
Zhilin Yang, Peng Qi, Saizheng Zhang, Yoshua Bengio, William~W. Cohen, Ruslan Salakhutdinov, and Christopher~D. Manning.
\newblock {{HotpotQA}}: {{A Dataset}} for {{Diverse}}, {{Explainable Multi-hop Question Answering}}, September 2018.

\bibitem[Yu et~al.(2025{\natexlab{a}})Yu, Zhang, Zhu, Yuan, Zuo, Yue, Dai, Fan, Liu, Liu, Liu, Lin, Lin, Ma, Sheng, Tong, Zhang, Zhang, Zhang, Zhu, Zhu, Chen, Chen, Wang, Yu, Song, Wei, Zhou, Liu, Ma, Zhang, Yan, Qiao, Wu, and Wang]{DAPOOpenSource2025}
Qiying Yu, Zheng Zhang, Ruofei Zhu, Yufeng Yuan, Xiaochen Zuo, Yu~Yue, Weinan Dai, Tiantian Fan, Gaohong Liu, Lingjun Liu, Xin Liu, Haibin Lin, Zhiqi Lin, Bole Ma, Guangming Sheng, Yuxuan Tong, Chi Zhang, Mofan Zhang, Wang Zhang, Hang Zhu, Jinhua Zhu, Jiaze Chen, Jiangjie Chen, Chengyi Wang, Hongli Yu, Yuxuan Song, Xiangpeng Wei, Hao Zhou, Jingjing Liu, Wei-Ying Ma, Ya-Qin Zhang, Lin Yan, Mu~Qiao, Yonghui Wu, and Mingxuan Wang.
\newblock {{DAPO}}: {{An Open-Source LLM Reinforcement Learning System}} at {{Scale}}, May 2025{\natexlab{a}}.

\bibitem[Yu et~al.(2025{\natexlab{b}})Yu, Ji, Wang, Yao, Wang, Cui, Yuan, Ding, Yao, Liu, Sun, and Chua]{RLPRExtrapolating2025}
Tianyu Yu, Bo~Ji, Shouli Wang, Shu Yao, Zefan Wang, Ganqu Cui, Lifan Yuan, Ning Ding, Yuan Yao, Zhiyuan Liu, Maosong Sun, and Tat-Seng Chua.
\newblock {{RLPR}}: {{Extrapolating RLVR}} to {{General Domains}} without {{Verifiers}}, June 2025{\natexlab{b}}.

\bibitem[Zhang et~al.(2025{\natexlab{a}})Zhang, Li, Hua, Lv, Ding, Qi, and Zhou]{OPENPRMBUILDING2025}
Kaiyan Zhang, Haoxin Li, Ermo Hua, Xingtai Lv, Ning Ding, Biqing Qi, and Bowen Zhou.
\newblock {{OPENPRM}}: {{BUILDING OPEN-DOMAIN PROCESS- BASED REWARD MODELS WITH PREFERENCE TREES}}.
\newblock 2025{\natexlab{a}}.

\bibitem[Zhang et~al.(2025{\natexlab{b}})Zhang, Zuo, He, Sun, Liu, Jiang, Fan, Tian, Jia, Li, Fu, Lv, Zhang, Zeng, Qu, Li, Wang, Wang, Long, Liu, Xu, Ma, Zhu, Hua, Liu, Li, Chen, Qu, Li, Chen, Yuan, Gao, Li, Ma, Cui, Liu, Qi, Ding, and Zhou]{SurveyReinforcement2025}
Kaiyan Zhang, Yuxin Zuo, Bingxiang He, Youbang Sun, Runze Liu, Che Jiang, Yuchen Fan, Kai Tian, Guoli Jia, Pengfei Li, Yu~Fu, Xingtai Lv, Yuchen Zhang, Sihang Zeng, Shang Qu, Haozhan Li, Shijie Wang, Yuru Wang, Xinwei Long, Fangfu Liu, Xiang Xu, Jiaze Ma, Xuekai Zhu, Ermo Hua, Yihao Liu, Zonglin Li, Huayu Chen, Xiaoye Qu, Yafu Li, Weize Chen, Zhenzhao Yuan, Junqi Gao, Dong Li, Zhiyuan Ma, Ganqu Cui, Zhiyuan Liu, Biqing Qi, Ning Ding, and Bowen Zhou.
\newblock A {{Survey}} of {{Reinforcement Learning}} for {{Large Reasoning Models}}, September 2025{\natexlab{b}}.

\bibitem[Zhang et~al.()Zhang, Li, Long, Zhang, Lin, Yang, Xie, Yang, Liu, Lin, Huang, and Zhou]{Qwen3Embedding2025}
Yanzhao Zhang, Mingxin Li, Dingkun Long, Xin Zhang, Huan Lin, Baosong Yang, Pengjun Xie, An~Yang, Dayiheng Liu, Junyang Lin, Fei Huang, and Jingren Zhou.
\newblock Qwen3 {{Embedding}}: {{Advancing Text Embedding}} and {{Reranking Through Foundation Models}}.
\newblock URL \url{http://arxiv.org/abs/2506.05176}.

\bibitem[Zhang et~al.(2025{\natexlab{c}})Zhang, Lu, Hu, Fu, Wen, Zhang, Liu, Jiang, Chen, Tang, Ding, Chen, Yang, Zhang, Gao, and Wang]{R1RewardTraining2025}
Yi-Fan Zhang, Xingyu Lu, Xiao Hu, Chaoyou Fu, Bin Wen, Tianke Zhang, Changyi Liu, Kaiyu Jiang, Kaibing Chen, Kaiyu Tang, Haojie Ding, Jiankang Chen, Fan Yang, Zhang Zhang, Tingting Gao, and Liang Wang.
\newblock R1-{{Reward}}: {{Training Multimodal Reward Model Through Stable Reinforcement Learning}}, May 2025{\natexlab{c}}.

\bibitem[Zhao et~al.(2025)Zhao, Huang, Hu, Wang, Mao, Zhang, Zhang, Jiang, Wu, Ai, Wang, Zhou, and Chen]{SWIFTScalable2025}
Yuze Zhao, Jintao Huang, Jinghan Hu, Xingjun Wang, Yunlin Mao, Daoze Zhang, Hong Zhang, Zeyinzi Jiang, Zhikai Wu, Baole Ai, Ang Wang, Wenmeng Zhou, and Yingda Chen.
\newblock {{SWIFT}}:{{A Scalable lightWeight Infrastructure}} for {{Fine-Tuning}}, May 2025.

\bibitem[Zheng et~al.(2023)Zheng, Chiang, Sheng, Zhuang, Wu, Zhuang, Lin, Li, Li, Xing, Zhang, Gonzalez, and Stoica]{JudgingLLMasaJudge2023}
Lianmin Zheng, Wei-Lin Chiang, Ying Sheng, Siyuan Zhuang, Zhanghao Wu, Yonghao Zhuang, Zi~Lin, Zhuohan Li, Dacheng Li, Eric~P. Xing, Hao Zhang, Joseph~E. Gonzalez, and Ion Stoica.
\newblock Judging {{LLM-as-a-Judge}} with {{MT-Bench}} and {{Chatbot Arena}}, December 2023.

\bibitem[Zheng et~al.(2024)Zheng, Yin, Xie, Sun, Huang, Yu, Cao, Kozyrakis, Stoica, Gonzalez, Barrett, and Sheng]{SGLangEfficient2024}
Lianmin Zheng, Liangsheng Yin, Zhiqiang Xie, Chuyue Sun, Jeff Huang, Cody~Hao Yu, Shiyi Cao, Christos Kozyrakis, Ion Stoica, Joseph~E. Gonzalez, Clark Barrett, and Ying Sheng.
\newblock {{SGLang}}: {{Efficient Execution}} of {{Structured Language Model Programs}}, June 2024.

\bibitem[Zhong et~al.(2025)Zhong, Shen, Li, Gao, Lu, Chen, Zhang, Zhou, Gu, and Zou]{ComprehensiveSurvey2025}
Jialun Zhong, Wei Shen, Yanzeng Li, Songyang Gao, Hua Lu, Yicheng Chen, Yang Zhang, Wei Zhou, Jinjie Gu, and Lei Zou.
\newblock A {{Comprehensive Survey}} of {{Reward Models}}: {{Taxonomy}}, {{Applications}}, {{Challenges}}, and {{Future}}, April 2025.

\bibitem[Zou et~al.(2025)Zou, Yang, Gu, Qiu, Shen, He, and Wang]{ReasonFluxPRMTrajectoryAware2025}
Jiaru Zou, Ling Yang, Jingwen Gu, Jiahao Qiu, Ke~Shen, Jingrui He, and Mengdi Wang.
\newblock {{ReasonFlux-PRM}}: {{Trajectory-Aware PRMs}} for {{Long Chain-of-Thought Reasoning}} in {{LLMs}}, June 2025.

\end{thebibliography}
